%% file: acl_latex.tex
\pdfoutput=1

\documentclass[11pt]{article}

\usepackage[table]{xcolor}
\usepackage[]{acl}

\usepackage{times}
\usepackage{latexsym}

\usepackage[T1]{fontenc}

\usepackage[utf8]{inputenc}

\usepackage{microtype}

\usepackage{inconsolata}

\usepackage{graphicx}
\usepackage{wrapfig}
\usepackage{subfigure}
\usepackage{booktabs}
\usepackage{longtable}

\usepackage{float}
\usepackage{tabularx}

\usepackage[most]{tcolorbox}
\usepackage{pgfplots}
\usepackage{hyperref}
\usepackage{xspace}
\usepackage{cleveref}
\usepackage{enumitem}
\usepackage{pgf-pie}
\usepackage{caption}
\usepackage{afterpage}
\usepackage{makecell}
\usepackage{pifont}
\usepackage{marvosym}

\definecolor{lightgreen}{RGB}{230,255,230}
\definecolor{lightblue}{RGB}{230,230,255}
\definecolor{lightred}{RGB}{255,230,230}
\definecolor{lightgrey}{RGB}{245,245,245}

\newcommand{\cmark}{\textcolor{green!35!black}{\ding{51}}}
\newcommand{\xmark}{\ding{55}}
\definecolor{YaleBlue}{RGB}{16, 42, 86}
\definecolor{NYUPurple}{RGB}{134, 1, 175}
\definecolor{NotreDameYellow}{RGB}{205, 184, 115}
\newcommand{\github}{\raisebox{-1.5pt}{\includegraphics[height=1.05em]{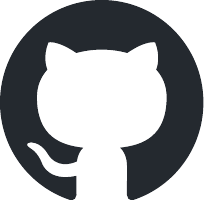}}\xspace}

\title{\ours: Benchmarking Foundation Models on University-Level\\Physics Problem Solving}

\author{
Kaiyue Feng$^{\hspace{.1em}{\textcolor{NYUPurple}{\boldsymbol{N}}}}$\thanks{Equal contributions.
Correspondence: \texttt{kf2365@nyu.edu, \{yilun.zhao, john.sous, arman.cohan\}@yale.edu}. Full author contributions in Appendix~\ref{sec:contributions}.} \quad
Yilun Zhao$^{\hspace{.1em}\textcolor{YaleBlue}{\boldsymbol{Y}}*}$ \quad
Yixin Liu$^{\hspace{.1em}\textcolor{YaleBlue}{\boldsymbol{Y}}}$ \quad
Tianyu Yang$^{\hspace{.1em}\textcolor{NotreDameYellow}{\boldsymbol{ND}}}$ \\
\bf{
Chen Zhao$^{\hspace{.1em}{\textcolor{NYUPurple}{\boldsymbol{N}}}}$ \quad
John Sous$^{\hspace{.1em}\textcolor{YaleBlue}{\boldsymbol{Y}}}$ \quad
Arman Cohan$^{\hspace{.1em}\textcolor{YaleBlue}{\boldsymbol{Y}}}$}
\vspace{5pt}\\
$^{\textcolor{YaleBlue}{\boldsymbol{Y}}}$Yale University \quad
$^{\textcolor{NYUPurple}{\boldsymbol{N}}}$New York University \quad
$^{\textcolor{NotreDameYellow}{\boldsymbol{ND}}}$University of Notre Dame
\vspace{5pt}
\\
\github ~~~\url{https://github.com/yale-nlp/physics-verified}
}

\newcommand{\ours}{\textsc{Physics}\xspace}
\newcommand{\nexample}{1,034\xspace}
\newcommand{\nmodel}{53\xspace}
\newcommand{\nsubject}{six\xspace}

\renewcommand{\cite}{\citep}

\pgfplotsset{compat=1.18}

\begin{document}

\maketitle

\begingroup
\clubpenalty=10000
\widowpenalty=10000
\displaywidowpenalty=10000

\input{main/0-abstract}
\input{main/1-intro}

\input{main/5-related_works}
\input{main/2-dataset}
\input{main/3-experiment}

\input{main/4-analysis}

\input{main/6-conclusion}

\par
\endgroup

\bibliography{custom}

\onecolumn
\input{main/Appendix}

\end{document}

%% file: main/0-abstract.tex
\begin{abstract}
\looseness=-1

We introduce \ours, a benchmark for evaluating foundation models on advanced physics problem solving. It contains \nexample annotated, open-ended problems from PhD-qualifying exams across six core physics domains. Solving these problems requires applying domain-specific knowledge to physical setups and carrying out multi-step mathematical reasoning. Each problem includes a worked solution and independently checked reference answers. An LLM-based grading protocol evaluates individual answers and complete multi-part problems, supporting numeric, symbolic, and text responses. We evaluate \nmodel proprietary and open-weight models. On test problems without figures, high accuracy is concentrated in a small group of thinking models, with the strongest open-weight models approaching proprietary performance. Most open-weight models still solve fewer than half of these problems. Our analysis of Qwen3.5-9B reveals failures in applying physics knowledge and interpreting figures, with misread figures affecting several subquestions.

\end{abstract}

%% file: main/1-intro.tex
\section{Introduction}\label{Introduction}

\input{figures_tex/example_page1}
Recent advances in foundation models have driven growing interest in mathematical reasoning benchmarks \citep{DBLP:conf/emnlp/ChenYKLWMXWX23, DBLP:conf/iclr/FanMWHBLPWB25, DBLP:conf/acl/LiuZQDFZZZLC24, DBLP:journals/corr/abs-2411-04872}. While mathematics underpins logical reasoning, we explore applications requiring multi-step reasoning and domain-specific knowledge. Physics, a core natural science, provides a mathematical framework for modeling \citep{jackson1998, sakurai2020, pathria2011}, spanning both deterministic classical mechanics and probabilistic quantum physics. Its complexity and multi-step reasoning demands make it an ideal domain for evaluating models on advanced problem solving.

\begin{figure*}[t]
    \centering
    \includegraphics[width=\textwidth]{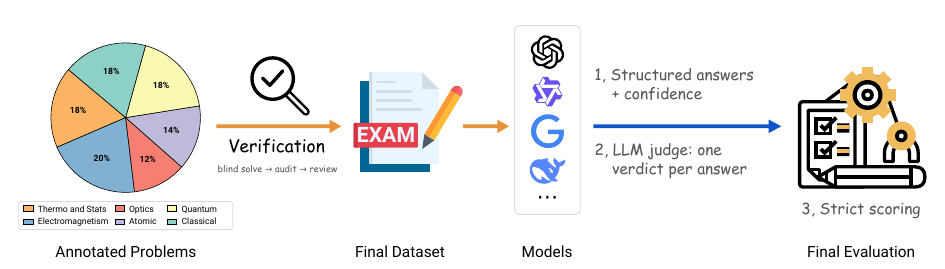}
    \caption{Overview of \ours. The pie chart uses the domain proportions of the final 1,034 problems. Annotated PhD-qualifying-exam problems pass through blind solving, auditing, and independent review (\S\ref{sec:verification}), then models produce structured responses (\S\ref{Experiment Setup}) for per-answer LLM grading (\S\ref{Automated Evaluation System}).}
    \label{fig:overall-process}
\end{figure*}

We present \ours, a benchmark designed to assess foundation models' physics problem-solving abilities. Unlike existing datasets that primarily consist of multiple-choice questions or focus on primary- to high school-level problems~\cite{DBLP:conf/aclnut/WelblLG17,DBLP:conf/nips/LuMX0CZTCK22,DBLP:conf/acl/YueZNW0T0Y000CN25, DBLP:conf/nips/HendrycksBKABTS21,DBLP:journals/corr/abs-2110-14168}, \ours is constructed from open-ended physics problems drawn from PhD-qualifying exams\footnotemark.
In contrast to multiple-choice questions that may allow models to exploit shortcuts or rely on answer recognition~\cite{DBLP:conf/iclr/Zheng0M0H24, DBLP:journals/corr/abs-2406-07545}, \ours is composed of problems that demand theoretical understanding, complex derivations, and the ability to integrate professional knowledge (\Cref{question_example}).

\footnotetext{Physics PhD-qualifying exams cover the core areas of physics that every PhD student must master to qualify as a physicist. Traditionally, students need to pass this challenging exam before advancing to the research phase of their PhD program. These exams are typically difficult, with passing thresholds ranging from 40\% to 70\%.}\ours spans \nsubject core physics domains: classical mechanics, quantum mechanics, thermodynamics and statistical mechanics, electromagnetism, atomic physics, and optics. Each problem comes with a set of reference answers, each labeled with its subquestion and answer type. Models answer in the response format of HLE \citep{DBLP:journals/corr/abs-2501-14249}, extended to problems with several answers. An LLM judge (GLM-5.3-Flash) receives the question, the response, and the reference answers in one call and issues one verdict per answer, following written rules for numeric, symbolic, and text equivalence. Strict accuracy, the fraction of problems for which every reference answer is judged correct, is the main metric. \Cref{fig:overall-process} shows the pipeline from problem collection through evaluation.

We evaluate \nmodel proprietary and open-weight models by strict accuracy on the 602 test problems without figures (\Cref{tab:llm-accuracy}). Leading open-weight models approach proprietary performance: GPT-6 Astra achieves 90.4\%, and Qwen3.8-27B reaches 88.5\%. High accuracy is concentrated in a small group of models, while most open-weight models solve fewer than half of these problems. The highest-ranked models all use thinking (\S\ref{sec:reasoning}). All open-weight multimodal models score lower on test problems with figures than on those without, whereas the proprietary models show no average decrease (\S\ref{sec:figures}). To characterize model errors, we analyze the wrong answers of Qwen3.5-9B, a mid-ranked thinking model, on all test problems. The analysis reveals failures in selecting physical models and interpreting figures, with figure misreadings often affecting several subquestions (\S\ref{sec:error-analysis}).

We summarize our contributions as follows:
\begin{itemize} [leftmargin=*]
\itemsep0em
\item
We present 1,034 annotated problems from physics PhD-qualifying exams in six domains. The main audit corrected a reference answer in 245 released problems; the final review removed 75 problems with confirmed defects (\S\ref{sec:verification}).

\item
We extend HLE-style LLM grading to multi-answer problems: each reference answer is labeled with its subquestion and type, and the judge, validated by a synthetic self-test and a second judge, returns one verdict per answer.

\item
We evaluate 53 proprietary and open-weight models, compare performance across thinking status, figures, and physics domains, and analyze the errors of Qwen3.5-9B.

\end{itemize}

\input{figures_tex/generated/dataset_stats}

%% file: figures_tex/example_page1.tex
\begin{figure}[!t]
    \centering
    \begingroup
    \tcbset{boxrule=0pt, before skip=0pt, after skip=0pt, top=1mm, bottom=1mm}
    \renewcommand{\baselinestretch}{1}\selectfont
    \fontsize{6pt}{8pt}\selectfont
    \setlength{\baselineskip}{8pt}
    \setlength{\abovedisplayskip}{3pt}
    \setlength{\belowdisplayskip}{3pt}
    \setlength{\abovedisplayshortskip}{2pt}
    \setlength{\belowdisplayshortskip}{2pt}

    \begin{tcolorbox}[colback=blue!5, sharp corners=south]
        \textcolor{blue}{\textbf{Problem:}}

        \setlength{\intextsep}{2pt}
        \setlength{\columnsep}{4pt}
        \begin{wrapfigure}[7]{r}{0.217\linewidth}
            \centering
            \includegraphics[width=\linewidth, trim=66.5bp 9.5bp 63bp 24.5bp, clip]{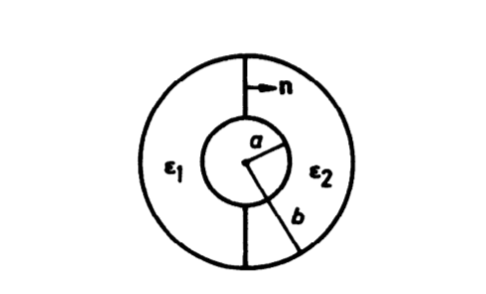}
        \end{wrapfigure}

        {\obeylines
One half of the region between the plates of a spherical capacitor of inner and outer radii $a$ and $b$ is filled with a linear isotropic dielectric of permittivity $\varepsilon_1$ and the other half has permittivity $\varepsilon_2$, as shown in Fig. 1.29. If the inner plate has total charge $Q$ and the outer plate has total charge $-Q$, find:
   
(a) the electric displacements $D_1$ and $D_2$ in the region of $\varepsilon_1$ and $\varepsilon_2$;  
(b) the electric fields in $\varepsilon_1$ and $\varepsilon_2$;  
(c) the total capacitance of this system.
        }

    \end{tcolorbox}

    \begin{tcolorbox}[colback=red!5, sharp corners=all]
        \textcolor{red}{\textbf{Solution:}}

The boundary conditions at the interface are $E_{1t} = E_{2t}, \quad D_{1n} = D_{2n}$. With $E_1 = E_2 = Ar/r^3$, a concentric Gaussian surface ($a < r < b$) gives $\oint D \cdot dS = Q$, so
\[
2\pi(\varepsilon_1 + \varepsilon_2)A = Q, \quad A = \frac{Q}{2\pi(\varepsilon_1 + \varepsilon_2)}.
\]

The electric fields and displacements are therefore
\[
\begin{aligned}
E_1 = E_2 &= \frac{Qr}{2\pi(\epsilon_1 + \epsilon_2) r^3}, \\
D_i &= \frac{\epsilon_i Qr}{2\pi(\epsilon_1 + \epsilon_2) r^3}\quad (i=1,2).
\end{aligned}
\]

Both semispherical regions have the same potential difference,
\[
V_{ab} = - \int_{b}^{a} \frac{A}{r^2} \, dr = \frac{A(b-a)}{ab}
\]

The charge on each inner hemisphere is $Q_i = 2\pi\epsilon_i A$, giving
\[
C_i = \frac{Q_i}{V_{ab}} = \frac{2\pi\epsilon_i ab}{b-a}\quad (i=1,2).
\]
The two semispherical capacitors are in parallel, so
\[
C = C_1 + C_2 = \frac{2\pi (\epsilon_1 + \epsilon_2) ab}{b-a} .
\]
    \end{tcolorbox}

    \begin{tcolorbox}[colback=green!5, sharp corners=north]
        \textcolor{green!50!black}{\textbf{Reference answers:}}

        \renewcommand{\arraystretch}{1.4}
        \begin{tabular}{@{}lll@{}}
        \textbf{Label} & \textbf{Answer} & \textbf{Type} \\
(a) & $D_1 = \frac{\epsilon_1 Qr}{2\pi(\epsilon_1 + \epsilon_2) r^3}$ & symbolic \\
(a) & $D_2 = \frac{\epsilon_2 Qr}{2\pi(\epsilon_1 + \epsilon_2) r^3}$ & symbolic \\
(b) & $E_1 = E_2 = \frac{Qr}{2\pi(\epsilon_1 + \epsilon_2) r^3}$ & symbolic \\
(c) & $C = \frac{2\pi (\epsilon_1 + \epsilon_2) ab}{b-a}$ & symbolic \\
        \end{tabular}
    \end{tcolorbox}
    \endgroup
\vspace{10pt}
\caption{An example problem from \ours (Electromagnetism, three subquestions and four reference answers). The solution shows the derivation; answer labels and types support automated grading.}
\label{question_example}
\end{figure}

%% file: figures_tex/generated/dataset_stats.tex
\begin{figure*}[t]
\centering
\begin{minipage}[c]{0.47\textwidth}
\centering
\small
\setlength{\tabcolsep}{3pt}
\renewcommand{\arraystretch}{0.96}
\begin{tabular}{@{}lr@{}}
\toprule
\textbf{Statistics} & \textbf{Value} \\
\midrule
Total problems & 1,034 \\
\quad With figures & 248 \\
\quad Validation : Test & 240 : 794 \\
Total answers & 2,559 \\
Answers per problem (mean) & 2.47 \\
\midrule
\multicolumn{2}{l}{\textbf{Answer types}} \\
\quad Numeric & 789 \\
\quad Symbolic & 1,547 \\
\quad Text & 223 \\
\midrule
\multicolumn{2}{l}{\textbf{Domain distribution}} \\
\quad Atomic Physics & 145 \\
\quad Electromagnetism & 211 \\
\quad Classical Mechanics & 188 \\
\quad Optics & 120 \\
\quad Quantum Mechanics & 188 \\
\quad Thermodynamics and Statistical Mechanics & 182 \\
\midrule
Avg. question length (words) & 82.9 \\
Avg. solution length (words) & 231.1 \\
\bottomrule
\end{tabular}
\end{minipage}
\hfill
\begin{minipage}[c]{0.50\textwidth}
\centering
\includegraphics[width=\linewidth]{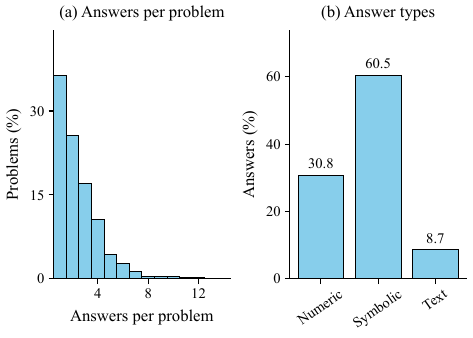}
\end{minipage}
\caption{Dataset statistics of \ours. The panels show the distribution of answers per problem (left; percentage of problems) and answer types (right; percentage of answers). Text lengths count whitespace-delimited tokens in the raw question and solution strings, including LaTeX expressions.}
\label{tab:dataset_composition}
\label{fig:dataset_reasoning}
\end{figure*}

%% file: main/5-related_works.tex
\section{Related Work}

\subsection{Scientific Reasoning Benchmarks}

Scientific reasoning benchmarks cover problems ranging from school-level science and mathematics~\citep{DBLP:conf/nips/LuMX0CZTCK22,DBLP:journals/corr/abs-2110-14168} to expert and research-level tasks~\citep{DBLP:journals/corr/abs-2311-12022,DBLP:journals/corr/abs-2501-14249,DBLP:journals/corr/abs-2411-04872}. Physics evaluation includes both multidisciplinary collections~\citep{DBLP:conf/cvpr/YueNZ0LZSJRSWYY24,DBLP:conf/aaai/SunHZMSC0024} and dedicated benchmarks built around coursework, competitions, and research problems~\citep{DBLP:conf/icml/XuXXCYZDY025,DBLP:conf/nips/QiuGSSCWLYZHWTC25,DBLP:journals/corr/abs-2509-26574}. These benchmarks examine scientific knowledge and mathematical reasoning, with multimodal tasks also testing figure interpretation~\citep{DBLP:conf/nips/XiangLZHLQHCYHX25}.

Reference quality is central to reliable scientific evaluation. Studies of existing benchmarks show that reference errors can affect model rankings, motivating systematic verification of questions and answers~\citep{DBLP:conf/naacl/GemaLHDMSHZDMBMHKKM25,DBLP:journals/corr/abs-2602-13964}. \ours focuses on open-ended PhD-qualifying-exam problems across six core physics domains, with worked solutions, verified reference answers, and separate grading of each answer in multi-part problems. \Cref{tab:dataset_comparison} compares its task scope and evaluation design with related benchmarks.

\subsection{Grading Open-Ended Answers}

\input{figures_tex/dataset_compare}

LLM judges are common for open-ended evaluation~\citep{DBLP:conf/nips/ZhengC00WZL0LXZ23,DBLP:conf/emnlp/LiuIXWXZ23}. Deciding whether a scientific answer is correct is a different task. Judges tend to accept answers from stronger models even when they are wrong~\citep{DBLP:journals/corr/abs-2409-04168}, and in physics assessment their validity depends on whether the answer can be checked against a reference~\citep{DBLP:journals/corr/abs-2603-14732}. JudgeBench~\citep{DBLP:conf/iclr/TanZMTC0PS25} tests judges on objective correctness rather than preferences. For scientific answers, PEDANTS~\citep{DBLP:conf/emnlp/LiMNLB24} and SCI-Verifier~\citep{DBLP:journals/corr/abs-2509-24285} study answer equivalence, while UGPhysics~\citep{DBLP:conf/icml/XuXXCYZDY025} combines typed rules with a model fallback for physics grading. As in HLE, the \ours judge compares a response with the reference answers and does not solve the problem itself.

%% file: figures_tex/dataset_compare.tex
\begin{table*}[t]
    \centering
    \resizebox{\textwidth}{!}{\begin{tabular}{lccccc}
        \toprule
        \textbf{Benchmark} & \textbf{Multi-modal} & \textbf{Size} & \textbf{Level} & \textbf{Question Type} & \textbf{Evaluation} \\
        \midrule
        JEEBench~\cite{DBLP:conf/emnlp/AroraSM23} & \xmark & 515 & CEE & OE, MC & Rule-Based \\
        MATH~\cite{DBLP:conf/nips/HendrycksBKABTS21} & \xmark & 12,500 & K12--Comp & OE & Rule-Based \\
        HARDMath~\cite{DBLP:conf/iclr/FanMWHBLPWB25} & \xmark & 1,466 & Graduate & OE & Rule + Model \\
        GSM8K~\cite{DBLP:journals/corr/abs-2110-14168} & \xmark & 8,500 & K8 & OE & Rule-Based \\
        GPQA~\cite{DBLP:journals/corr/abs-2311-12022} & \xmark & 448 & Graduate & MC & Rule-Based \\
        SciQ~\cite{DBLP:conf/aclnut/WelblLG17} & \xmark & 13,679 & K4--K8 & MC, OE & Rule-Based \\
        SciEval~\cite{DBLP:conf/aaai/SunHZMSC0024} & \xmark & -- & -- & OE, MC & Rule + Manual \\
        PutnamBench~\cite{DBLP:conf/nips/TsoukalasLJXDJT24} & \xmark & 1,692 & University & OE & Proof assistant \\
        PHYBench~\cite{DBLP:conf/nips/QiuGSSCWLYZHWTC25} & \xmark & 500 & HS--Comp & OE & Rule-Based \\
        UGPhysics~\cite{DBLP:conf/icml/XuXXCYZDY025} & \xmark & 5,520 & University & OE, MC & Rule + Model \\
        SciBench~\cite{DBLP:conf/icml/WangHL0ZSLZS024} & \cmark & 869 & University & OE & Rule-Based \\
        MMMU~\cite{DBLP:conf/cvpr/YueNZ0LZSJRSWYY24} & \cmark & 11,550 & University & OE, MC & Rule-Based \\
        MMMU-Pro~\cite{DBLP:conf/acl/YueZNW0T0Y000CN25} & \cmark & 1,730 & University & MC & Rule-Based \\
        MMVU~\cite{DBLP:conf/cvpr/00010XHGLHCLXWS25} & \cmark & 3,000 & University & OE, MC & Rule + Model \\
        ScienceQA~\cite{DBLP:conf/nips/LuMX0CZTCK22} & \cmark & 21,208 & K1--K12 & MC & Rule-Based \\
        OlympiadBench~\cite{DBLP:conf/acl/HeLBHTSHHHZLQL024} & \cmark & 8,476 & CEE--Comp & OE & Rule + Manual \\
        PhysReason~\cite{DBLP:conf/acl/ZhangDWHJFSZ025} & \cmark & 1,200 & CEE--Comp & OE & LLM Judge \\
        SeePhys~\cite{DBLP:conf/nips/XiangLZHLQHCYHX25} & \cmark & 2,000 & K12--PhD & OE & Rule + Model \\
        HLE~\cite{DBLP:journals/corr/abs-2501-14249} & \cmark & 2,500 & Expert & MC, OE & LLM Judge \\
        \rowcolor{gray!20} \ours & \cmark & 1,034 & PhD & OE & LLM Judge (per answer) \\
        \bottomrule
    \end{tabular}
    }
    \caption{Comparison of \ours with related benchmarks. For \textbf{Level},
K1--K12: Elementary through High School, K4--K8: Grades 4--8, K8: up to Grade 8, HS: High School, CEE: University Entrance Examination, Comp: Competition, University: Undergraduate, PhD: PhD Qualifying Exams. For \textbf{Question Type}, OE: Open-ended Questions, MC: Multiple-choice Questions. Sizes use the sets and counting units indicated by the cited sources; multi-subject benchmarks are not restricted to their physics subsets.}

\label{tab:dataset_comparison}
\end{table*}

%% file: main/2-dataset.tex
\section{\ours Benchmark}
\label{Data collection}

\ours has the following key features:
(1) \textbf{Six Physics Domains}: problems span six core fields of physics (\S\ref{sec:problem-selection}).
(2) \textbf{Multi-Step Reasoning}: solutions combine several derivation steps.
(3) \textbf{Individually Gradable Answers}: each reference answer is labeled with its subquestion and typed as numeric, symbolic, or text, allowing a judge to issue one verdict per answer.
(4) \textbf{Verification}: every problem was solved without its reference, audited against it, and reviewed separately; the main audit corrected a reference answer in 245 released problems, and the final review removed 75 problems with confirmed defects (\S\ref{sec:verification}).

\subsection{Problem Selection}
\label{sec:problem-selection}
The problems come from publicly available sources and cover classical mechanics, quantum mechanics, thermodynamics and statistical mechanics, electromagnetism, atomic physics, and optics. An example is shown in \Cref{question_example}.

\subsection{Data Annotation}
\label{Data Annotation}
A team of seven physics-trained annotators carried out the annotation (biographies in \Cref{tab:annotators}). Each annotator passed a qualification test before contributing and checked that the selected problems, drawn from major textbooks and reputable online resources, adhered to copyright and licensing regulations. Annotators formatted each problem statement in \LaTeX\ with consistent notation and variable usage, recorded its conditions and assumptions, included any figure that shows the physical setup, and wrote a step-by-step solution with derivations and final answers. The authors then reviewed every annotation and resolved ambiguities with the annotators, and a second annotator checked each problem for accuracy and consistency.

\subsection{Verification Pipeline}
\label{sec:verification}

Verification involved four activities carried out by LLM agents (Claude Opus 5, GPT-5.6 Sol, GPT-6 Astra) under fixed written rules, with independent review (details in \Cref{sec:verification-details}).

\paragraph{Results-only scope.} We kept subquestions that ask for a final result (a value, expression, or short statement) and deleted requests for proofs, explanations, and drawings without changing the retained wording. Across the 1,297 annotated problems, 2,730 subquestions were marked for retention and 752 for deletion. A subquestion can request several separately graded answers.

\paragraph{Blind solving and audit.} A solver answered each problem without the reference, an auditor compared that answer with the reference and corrected errors established by independent derivation, and a separate reviewer with no earlier context checked each batch of up to five problems, including unchanged and removed ones. Repairs corrected answers and solutions while preserving retained question wording. Among the released problems, the main audit corrected reference answers in 245 and worked solutions only in 264 more.

\paragraph{Answer annotation.} Each answer was labeled with its subquestion and typed as numeric, symbolic, or text. Two independent passes produced identical labels and types for 1,064 of the 1,126 problems available at this stage (94.5\%), and an adjudicator decided the other 62.

\paragraph{Model-output checks.} We screened references that Qwen3.5-9B and Qwen3.8-27B were both judged wrong on while giving the same answer. Each lead was re-derived independently and each proposed error checked by a second verifier; 5 references were corrected. Scope, dependency, duplicate, and reference checks removed 188 problems in all, leaving 1,109 problems with 2,803 answers. After evaluation, a final review of answers missed by all five leading models removed 75 problems with confirmed defects, yielding 1,034 problems with 2,559 answers. Removal categories are reported in \Cref{sec:verification-details}.

\subsection{Data Statistics}

\Cref{tab:dataset_composition} presents the statistics of the dataset. The problems are split into 240 validation and 794 test problems; 248 (24.0\%) include figures. Problems have 2.47 answers on average; 60.5\% of answers are symbolic, 30.8\% numeric, and 8.7\% text. The average question length is 82.9 words and the average solution length is 231.1 words.

%% file: main/3-experiment.tex
\section{Experiment}
\label{Experiments}

Our experiments measure how accurately foundation models solve \ours problems using a common response format and an LLM-based \mbox{grading protocol}.

\subsection{Experimental Setup}
\label{Experiment Setup}

We evaluate 53 models on \ours: 4 proprietary and 49 open-weight (\Cref{tab:llm-accuracy}). The proprietary models (GPT-6 Astra, GPT-6 Sol, GPT-5.6 Sol, and GPT-6 Luna~\citep{gpt6_systemcard,gpt56_systemcard}) are queried through the OpenAI API with reasoning effort set to high. The open-weight models are served with vLLM~\citep{DBLP:conf/sosp/KwonLZ0ZY0ZS23} using model-specific sampling settings recorded in the released outputs; they include models from the DeepSeek~\citep{DBLP:journals/corr/abs-2606-19348,DBLP:journals/nature/GuoYZSWZXZMBZY025}, Qwen~\citep{qwen3.5,qwen38,DBLP:journals/corr/abs-2412-15115}, Gemma~\citep{DBLP:journals/corr/abs-2607-02770,DBLP:journals/corr/abs-2408-00118}, GLM~\citep{DBLP:journals/corr/abs-2602-15763}, Llama~\citep{llama4_2025,DBLP:journals/corr/abs-2407-21783}, Mistral~\citep{mistralsmall4_2026,DBLP:journals/corr/abs-2601-08584}, InternVL~\citep{DBLP:journals/corr/abs-2508-18265,DBLP:journals/corr/abs-2412-05271}, and Phi~\citep{DBLP:journals/corr/abs-2603-03975,DBLP:journals/corr/abs-2412-08905} families, among others.

The output budget is 65,536 tokens for thinking models and 8,192 for non-thinking models; checkpoints with smaller context windows use the remaining available space. The 27 text-only models are evaluated on the 786 problems without figures; the 26 multimodal models on all 1,034 problems. The validation split (240 problems) is intended for development; the test split (794 problems) is for reporting. \Cref{tab:llm-accuracy} is sorted by strict accuracy on the 602 test problems without figures.

Models answer in the response format of HLE~\citep{DBLP:journals/corr/abs-2501-14249}, extended to problems with several answers: an explanation, the final answers one per line in the order asked (each with its part label when the question has parts), and a confidence score. The model is not told how many answers are expected.

\input{figures_tex/cot_prompt}

\subsection{Evaluation Protocol}
\label{Automated Evaluation System}

\input{figures_tex/generated/main_results}

Each response is graded by GLM-5.3-Flash~\citep{DBLP:journals/corr/abs-2602-15763}, served with vLLM on our own GPUs, in one call per problem that returns a verdict for each reference answer (prompt in \Cref{sec:prompts}). The judge receives the question text, the model's final response (reasoning traces are stripped), and the list of reference answers; figures are not passed to the judge. For each reference answer the judge extracts the model's stated result and decides correctness. Symbolic answers must be algebraically equivalent; a missing or different factor, sign, or exponent counts as incorrect. Numeric answers must fall within 1\% relative difference of the reference or equal the reference when rounded to its significant figures, with equivalent units required; for order-of-magnitude estimates, the same order of magnitude is correct. Text answers must state the same conclusion; any contradiction is incorrect. A response that hedges between alternatives, addresses a different quantity, or gives no final answer is marked incorrect. An answer that matches the reference once a convention stated in the response is applied (for example, Gaussian instead of SI units) counts as correct; an unstated convention does not. An empty response, for example one that reached the token limit before the model finished reasoning, is scored as wrong on every answer without a judge call; truncated responses that contain visible text are judged like any other. The confidence score is extracted separately and defaults to 100; it does not affect the accuracy metrics.

\paragraph{Metrics.}
Strict accuracy is the fraction of problems for which every reference answer is judged correct; it is the main metric and the one reported in \Cref{tab:llm-accuracy}. Answer-level accuracy pools all answer verdicts across problems. Partial credit averages the within-problem fraction correct, weighting every problem equally. Both are reported \mbox{for every model in \Cref{tab:full-results}}.

\paragraph{Judge validation.}
On a synthetic self-test with known verdicts, GLM-5.3-Flash returned the expected verdict for 202 of 204 target answers and for all 481 unchanged answers in the same responses. A cross-check with DeepSeek-V4.1-Flash on 200 problems answered by Qwen3.5-9B (476 answers) showed 93.5\% per-answer agreement; of the 31 disagreements, 29 had GLM-5.3-Flash accepting an answer the other judge rejected. Four independent human reviewers, shown the two judges only as A and B, adjudicated 27 in favor of GLM-5.3-Flash and 4 in favor of DeepSeek-V4.1-Flash.

%% file: figures_tex/cot_prompt.tex
\begin{tcolorbox}[colback=gray!10,colframe=black!75,title={\small Response Format Prompt}]
\footnotesize
\setlength{\baselineskip}{1.2\baselineskip}
Your response should be in the following format:\\[4pt]
\textbf{Explanation:} \{your explanation for your answers\}\\[2pt]
\textbf{Answers:} \{your final answers, one per line, in the order the question asks for them; begin each line with the subquestion label, such as (a), when the question labels its parts\}\\[2pt]
\textbf{Confidence:} \{your confidence score between 0\% and 100\% for your answers\}
\end{tcolorbox}

%% file: figures_tex/generated/main_results.tex
\begin{table*}[p]
\centering
\small
\setlength{\tabcolsep}{2.5pt}
\renewcommand{\arraystretch}{1.0}
\begin{tabular*}{\textwidth}{@{\extracolsep{\fill}}l c r r r r r r r r >{\columncolor{gray!20}}r@{}}
\toprule
\textbf{Model} & \textbf{Think} & \textbf{AMO} & \textbf{E\&M} & \textbf{CM} & \textbf{Opt.} & \textbf{QM} & \textbf{Stats.} & \textbf{Val} & \textbf{Test (w/ fig.)} & \textbf{Test} \\
 &  & {\footnotesize (101)} & {\footnotesize (119)} & {\footnotesize (83)} & {\footnotesize (51)} & {\footnotesize (127)} & {\footnotesize (121)} & {\footnotesize (184)} & {\footnotesize (794)} & {\footnotesize (602)} \\
\midrule
\multicolumn{11}{c}{\textbf{Proprietary Models}} \\
GPT-6 Astra\vphantom{\cmark} & \cmark & 94.1 & 92.4 & 89.2 & 80.4 & 91.3 & 89.3 & 92.4 & 91.1 & 90.4 \\
GPT-6 Sol\vphantom{\cmark} & \cmark & 93.1 & 92.4 & 89.2 & 86.3 & 88.2 & 89.3 & 90.2 & 90.6 & 90.0 \\
GPT-5.6 Sol\vphantom{\cmark} & \cmark & 90.1 & 93.3 & 89.2 & 84.3 & 90.6 & 88.4 & 90.8 & 90.4 & 89.9 \\
GPT-6 Luna\vphantom{\cmark} & \cmark & 83.2 & 85.7 & 85.5 & 78.4 & 83.5 & 78.5 & 83.2 & 82.6 & 82.7 \\
\midrule
\multicolumn{11}{c}{\textbf{Open-Weight Models}} \\
Qwen3.8-27B\vphantom{\cmark} & \cmark & 87.1 & 86.6 & 91.6 & 86.3 & 88.2 & 90.9 & 87.5 & 87.2 & 88.5 \\
DeepSeek-V4.1-Flash\vphantom{\cmark} & \cmark & 86.1 & 87.4 & 91.6 & 86.3 & 87.4 & 90.1 & 91.8 & 87.0 & 88.2 \\
GLM-5.3-Flash\vphantom{\cmark} & \cmark & 85.1 & 85.7 & 90.4 & 82.4 & 85.0 & 84.3 & 88.6 & 85.3 & 85.5 \\
Qwen3.5-122B-A10B\vphantom{\cmark} & \cmark & 81.2 & 83.2 & 84.3 & 74.5 & 78.0 & 81.8 & 85.3 & 80.7 & 80.9 \\
gemma-4-31B-it\vphantom{\cmark} & \cmark & 84.2 & 72.3 & 80.7 & 74.5 & 72.4 & 81.0 & 83.2 & 76.8 & 77.4 \\
Qwen3.5-9B\vphantom{\cmark} & \cmark & 73.3 & 76.5 & 71.1 & 70.6 & 75.6 & 79.3 & 78.3 & 73.8 & 75.1 \\
Mistral-Small-4-119B-2603\vphantom{\cmark} & \cmark & 66.3 & 62.2 & 61.4 & 64.7 & 64.6 & 63.6 & 72.8 & 61.7 & 63.8 \\
DeepSeek-R1\smash{$^{\dagger}$}\vphantom{\cmark} & \cmark & 65.3 & 54.6 & 65.1 & 62.7 & 52.8 & 71.1 & 65.2 & -- & 61.5 \\
DeepSeek-R1-0528-Qwen3-8B\smash{$^{\dagger}$}\vphantom{\cmark} & \cmark & 52.5 & 39.5 & 42.2 & 43.1 & 38.6 & 57.9 & 56.0 & -- & 45.8 \\
DeepSeek-R1-Distill-Qwen-32B\smash{$^{\dagger}$}\vphantom{\cmark} & \cmark & 54.5 & 37.0 & 39.8 & 49.0 & 33.9 & 44.6 & 45.1 & -- & 42.2 \\
Intern-S1-mini\vphantom{\cmark} & \cmark & 44.6 & 31.9 & 42.2 & 47.1 & 25.2 & 43.8 & 40.8 & 34.4 & 37.7 \\
QwQ-32B-Preview\smash{$^{\dagger}$}\vphantom{\cmark} & \cmark & 45.5 & 30.3 & 28.9 & 39.2 & 25.2 & 49.6 & 34.2 & -- & 36.2 \\
gemma-4-E4B-it\vphantom{\cmark} & \cmark & 43.6 & 31.9 & 37.3 & 41.2 & 23.6 & 41.3 & 33.7 & 31.7 & 35.5 \\
Ministral-3-14B-Instruct-2512\vphantom{\cmark} & \xmark & 48.5 & 29.4 & 26.5 & 43.1 & 15.0 & 39.7 & 37.0 & 29.6 & 32.4 \\
Ministral-3-8B-Instruct-2512\vphantom{\cmark} & \xmark & 40.6 & 29.4 & 24.1 & 56.9 & 18.9 & 33.9 & 31.0 & 27.7 & 31.6 \\
Llama-4-Scout-17B-16E-Instruct\vphantom{\cmark} & \xmark & 46.5 & 26.1 & 21.7 & 43.1 & 16.5 & 37.2 & 35.9 & 28.2 & 30.6 \\
Qwen2.5-Math-72B-Instruct\smash{$^{\dagger}$}\vphantom{\cmark} & \xmark & 34.7 & 24.4 & 30.1 & 39.2 & 18.9 & 36.4 & 34.8 & -- & 29.4 \\
Phi-4\smash{$^{\dagger}$}\vphantom{\cmark} & \xmark & 41.6 & 29.4 & 22.9 & 37.3 & 15.7 & 30.6 & 33.2 & -- & 28.6 \\
InternVL3.5-38B\vphantom{\cmark} & \xmark & 43.6 & 21.8 & 27.7 & 33.3 & 12.6 & 26.4 & 29.9 & 24.6 & 26.2 \\
Llama-3.3-70B-Instruct\smash{$^{\dagger}$}\vphantom{\cmark} & \xmark & 32.7 & 22.7 & 18.1 & 47.1 & 15.0 & 31.4 & 29.9 & -- & 25.9 \\
Qwen2.5-72B-Instruct\smash{$^{\dagger}$}\vphantom{\cmark} & \xmark & 32.7 & 23.5 & 21.7 & 39.2 & 12.6 & 28.1 & 27.2 & -- & 24.8 \\
gemma-4-E2B-it\vphantom{\cmark} & \cmark & 29.7 & 23.5 & 21.7 & 27.5 & 17.3 & 21.5 & 19.6 & 20.2 & 22.9 \\
Mistral-Small-24B-Instruct-2501\smash{$^{\dagger}$}\vphantom{\cmark} & \xmark & 32.7 & 17.6 & 16.9 & 29.4 & 10.2 & 27.3 & 21.2 & -- & 21.4 \\
Qwen3.5-2B\vphantom{\cmark} & \cmark & 31.7 & 16.0 & 21.7 & 29.4 & 9.4 & 24.8 & 19.0 & 19.4 & 20.9 \\
Ministral-3-8B-Reasoning-2512\vphantom{\cmark} & \cmark & 22.8 & 19.3 & 21.7 & 31.4 & 8.7 & 28.1 & 20.7 & 18.8 & 20.8 \\
Qwen2.5-32B-Instruct\smash{$^{\dagger}$}\vphantom{\cmark} & \xmark & 26.7 & 17.6 & 15.7 & 29.4 & 12.6 & 20.7 & 22.3 & -- & 19.4 \\
Qwen2.5-14B-Instruct\smash{$^{\dagger}$}\vphantom{\cmark} & \xmark & 26.7 & 17.6 & 18.1 & 27.5 & 9.4 & 20.7 & 17.4 & -- & 18.9 \\
InternVL2.5-38B\vphantom{\cmark} & \xmark & 28.7 & 14.3 & 15.7 & 27.5 & 7.1 & 23.1 & 19.0 & 16.4 & 18.3 \\
Phi-4-Reasoning-Vision-15B\vphantom{\cmark} & \cmark & 23.8 & 16.0 & 10.8 & 23.5 & 6.3 & 19.0 & 17.9 & 14.6 & 15.8 \\
gemma-2-27b-it\smash{$^{\dagger}$}\vphantom{\cmark} & \xmark & 28.7 & 11.8 & 9.6 & 17.6 & 3.1 & 17.4 & 10.3 & -- & 14.1 \\
Qwen2.5-Math-7B-Instruct\smash{$^{\dagger}$}\vphantom{\cmark} & \xmark & 18.8 & 13.4 & 10.8 & 19.6 & 6.3 & 13.2 & 15.8 & -- & 13.0 \\
Qwen2.5-Math-7B\smash{$^{\dagger}$}\vphantom{\cmark} & \xmark & 10.9 & 11.8 & 7.2 & 25.5 & 5.5 & 17.4 & 10.3 & -- & 12.0 \\
Qwen2-VL-72B-Instruct-AWQ\vphantom{\cmark} & \xmark & 22.8 & 12.6 & 8.4 & 19.6 & 1.6 & 8.3 & 11.4 & 9.7 & 11.1 \\
DeepSeekMath-7B-RL\smash{$^{\dagger}$}\vphantom{\cmark} & \xmark & 20.8 & 8.4 & 7.2 & 23.5 & 3.1 & 11.6 & 6.0 & -- & 11.1 \\
Qwen2.5-7B-Instruct\smash{$^{\dagger}$}\vphantom{\cmark} & \xmark & 20.8 & 10.1 & 9.6 & 15.7 & 3.9 & 8.3 & 7.1 & -- & 10.6 \\
Aria\vphantom{\cmark} & \xmark & 18.8 & 5.9 & 7.2 & 21.6 & 5.5 & 10.7 & 5.4 & 9.2 & 10.5 \\
gemma-2-9b-it\smash{$^{\dagger}$}\vphantom{\cmark} & \xmark & 15.8 & 9.2 & 7.2 & 15.7 & 3.9 & 9.9 & 6.0 & -- & 9.6 \\
Yi-1.5-34B-Chat\smash{$^{\dagger}$}\vphantom{\cmark} & \xmark & 16.8 & 4.2 & 9.6 & 21.6 & 2.4 & 10.7 & 8.7 & -- & 9.5 \\
Llama-3.1-8B-Instruct\smash{$^{\dagger}$}\vphantom{\cmark} & \xmark & 16.8 & 7.6 & 9.6 & 19.6 & 2.4 & 7.4 & 7.6 & -- & 9.3 \\
InternLM3-8B-Instruct-AWQ\smash{$^{\dagger}$}\vphantom{\cmark} & \xmark & 16.8 & 8.4 & 9.6 & 7.8 & 3.1 & 7.4 & 11.4 & -- & 8.6 \\
Qwen2.5-Math-1.5B-Instruct\smash{$^{\dagger}$}\vphantom{\cmark} & \xmark & 10.9 & 8.4 & 9.6 & 11.8 & 2.4 & 9.9 & 6.0 & -- & 8.3 \\
Mathstral-7B-v0.1\smash{$^{\dagger}$}\vphantom{\cmark} & \xmark & 12.9 & 8.4 & 4.8 & 9.8 & 3.1 & 7.4 & 3.3 & -- & 7.5 \\
Pixtral-12B-2409\vphantom{\cmark} & \xmark & 10.9 & 8.4 & 4.8 & 13.7 & 1.6 & 5.0 & 8.2 & 5.4 & 6.6 \\
c4ai-command-r-v01\smash{$^{\dagger}$}\vphantom{\cmark} & \xmark & 9.9 & 5.0 & 3.6 & 7.8 & 0.8 & 5.0 & 2.7 & -- & 5.0 \\
Mistral-7B-Instruct-v0.3\smash{$^{\dagger}$}\vphantom{\cmark} & \xmark & 10.9 & 3.4 & 2.4 & 7.8 & 0.0 & 4.1 & 4.9 & -- & 4.3 \\
gemma-2-2b-it\smash{$^{\dagger}$}\vphantom{\cmark} & \xmark & 8.9 & 2.5 & 1.2 & 9.8 & 0.8 & 5.0 & 3.3 & -- & 4.2 \\
GLM-4-9B-Chat-hf\smash{$^{\dagger}$}\vphantom{\cmark} & \xmark & 9.9 & 3.4 & 3.6 & 5.9 & 0.0 & 2.5 & 4.3 & -- & 3.8 \\
DeepSeek-VL2-Small\vphantom{\cmark} & \xmark & 7.9 & 4.2 & 1.2 & 5.9 & 0.8 & 3.3 & 2.2 & 3.0 & 3.7 \\
ChatGLM3-6B\smash{$^{\dagger}$}\vphantom{\cmark} & \xmark & 4.0 & 1.7 & 0.0 & 3.9 & 0.8 & 2.5 & 3.3 & -- & 2.0 \\
\bottomrule
\end{tabular*}
\caption{Strict accuracy (\%). Domain columns and Test: 602 test problems without figures; Val: 184 validation problems without figures; Test (w/ fig.): all 794 test problems. Models are sorted by Test within groups; $^{\dagger}$ marks text-only models (-- for Test (w/ fig.)). \textbf{Domains:} AMO (Atomic Physics), E\&M (Electromagnetism), CM (Classical Mechanics), Opt. (Optics), QM (Quantum Mechanics), Stats. (Thermodynamics and Statistical Mechanics).}
\label{tab:llm-accuracy}
\end{table*}

%% file: main/4-analysis.tex
\section{Results and Analysis}
\label{sec:results}

We analyze overall model performance, differences across thinking status, figures, and physics domains, and the errors that lead to incorrect answers.

\subsection{Main Findings}

\Cref{tab:llm-accuracy} reports strict accuracy on \ours. Unless stated otherwise, the numbers below are strict accuracy on the 602 test problems without figures, which every model answered.

\paragraph{Leading open-weight models approach proprietary performance.}
GPT-6 Astra solves 90.4\% of problems, followed by GPT-6 Sol (90.0\%) and GPT-5.6 Sol (89.9\%). Qwen3.8-27B and DeepSeek-V4.1-Flash score 88.5\% and 88.2\%, respectively, placing the strongest open-weight models close to the proprietary leaders (\Cref{sec:appendix-ownership}). Qwen3.8-27B's earlier same-family counterpart at a similar size, Qwen2.5-32B-Instruct, solves 19.4\%. Qwen3.5-9B (75.1\%) also outperforms the larger Qwen2.5-72B-Instruct (24.8\%) (\Cref{sec:appendix-generations}).

\paragraph{Most open-weight models still struggle.}
Of the 49 open-weight models, 41 solve fewer than half of the problems. No non-thinking model exceeds 32.4\% (Ministral-3-14B-Instruct-2512), and the median of the 32 non-thinking models is 11.1\%.

\begin{figure}[t]
  \centering
  \includegraphics[width=\linewidth]{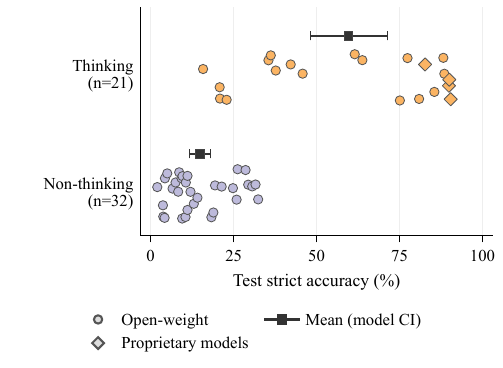}
  \caption{Strict accuracy on test problems without figures by thinking status. Circles denote open-weight models, diamonds proprietary models. Squares show group means with model-resampling 95\% intervals.}
  \label{fig:analysis-reasoning}
\end{figure}

\paragraph{Thinking models dominate the top of the ranking.}
\phantomsection
\label{sec:reasoning}
The 17 highest-ranked models all use thinking (\Cref{fig:analysis-reasoning}). Thinking models average 59.6\% against 14.8\% for non-thinking models, a gap of 44.9 points. Among related checkpoints, DeepSeek-R1-Distill-Qwen-32B reaches 42.2\% compared with 19.4\% for Qwen2.5-32B-Instruct, while Ministral-3-8B-Reasoning-2512 scores 20.8\% against 31.6\% for its instruct sibling. Qwen3.5-2B scores 20.9\%, compared with \mbox{GPT-6 Astra's 90.4\%}.

\begin{figure}[t]
  \centering
  \includegraphics[width=\linewidth]{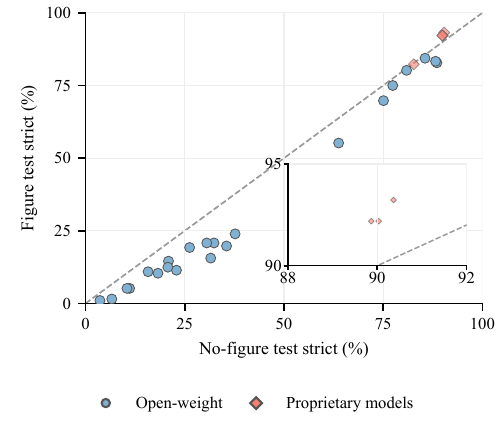}
  \caption{Strict accuracy on problems with versus without figures for each multimodal model. The dashed line denotes equal scores; the inset separates the overlapping proprietary-model points near the ceiling.}
  \label{fig:analysis-figures}
\end{figure}

\paragraph{Open-weight models score lower on problems with figures.}
\phantomsection
\label{sec:figures}
Every one of the 22 open-weight multimodal models scores lower on problems with figures (\Cref{fig:analysis-figures}), by 7.3 points on average and by 4.1 points among the seven models with at least 60\% accuracy without figures. The four OpenAI models score 1.7 points higher on average with figures. The open-weight gap persists after reweighting by domain (8.2 points). In the error analysis of Qwen3.5-9B, figure misreading is the second most common error type by answer count: incorrect interpretations of labels and geometric relations often propagate across several subquestions (\S\ref{sec:error-analysis}).

\paragraph{Domain difficulty varies with the scoring metric and model group.}
\phantomsection
\label{sec:domains}
Quantum Mechanics has the lowest mean strict accuracy (25.7\%). Its problems ask for 2.91 answers on average, against at most 2.41 in other domains, and strict scoring requires all of them. At answer level, Electromagnetism (38.8\%) and Classical Mechanics (39.3\%) are lowest. Among the ten strongest models, Optics has the lowest mean strict accuracy (80.4\%) and Classical Mechanics the highest (86.3\%), with Quantum Mechanics at 84.0\% (\Cref{sec:appendix-domains}).

\input{main/4b-errors}

%% file: main/4b-errors.tex
\subsection{Error Analysis}
\label{sec:error-analysis}

We analyzed the wrong answers of Qwen3.5-9B, a mid-ranked thinking model with released reasoning traces, on the 794 test problems. Of its 208 failed problems, 41 (19.7\%) have a length flag (reached the 65,536-token output budget) and are excluded from per-answer analysis. We selected 173 incorrect answers with clearly identifiable model errors from 118 problems and grouped the remaining 49 nontruncated failures as other cases. Below, we describe seven error types with their \mbox{answer-level frequencies}.

\paragraph{Lack of professional knowledge} (40 answers, 23.1\%, across 31 problems).
The model uses the wrong physical law or model, or leaves out an effect that changes the answer. In one electromagnetism problem, Qwen3.5-9B invents opposing bound surface-current sheets at the gap of a magnetized torus, producing a field with the \mbox{wrong sign (\Cref{appendix:knowledge})}.

\paragraph{Figure misreading} (35 answers, 20.2\%, across 14 problems).
The model reads a label or geometric relation in the figure wrongly. A misread figure often fails several parts of the same problem, and two circuit problems contribute 11 and 6 answers. In the appendix example (\Cref{appendix:figure}), the model swaps the two radii labeled in the figure.

\paragraph{Calculation errors} (28 answers, 16.2\%, across 23 problems).
In these errors the physical setup is correct and the algebra is not. In a problem about a dipole above a grounded plane, Qwen3.5-9B correctly derives the image-dipole interaction energy $U \propto -\tfrac{1}{2}\,\mathbf{p}\cdot\mathbf{E}$ but drops a negative sign when simplifying, selecting $\theta=90^{\circ}$ instead of $\theta=0^{\circ}$ or $180^{\circ}$ as the equilibrium orientation. Another example is in \Cref{appendix:calculation}.

\paragraph{Answer-form mismatch} (28 answers, 16.2\%, across 22 problems).
The model's physics is correct, but its final answer is qualitative where a formula is required, omits a requested quantity, or gives conflicting final answers (\Cref{appendix:answer-form}).

\paragraph{Non-existent assumptions} (19 answers, 11.0\%, across 18 problems).
The model introduces an approximation or simplification not warranted by the problem. In the appendix example (\Cref{appendix:assumption}), Qwen3.5-9B recognizes the relativistic Doppler formula but drops the Lorentz factor based on a guess about the textbook chapter.

\paragraph{Misunderstanding the question} (19 answers, 11.0\%, across 15 problems).
The model answers a different question than the one posed. In one quantum mechanics problem, it computes an expectation value when the problem asks for a measurement probability. The appendix example (\Cref{appendix:misunderstanding}) assigns the two binding energies to two different sites instead of two states per site.

Incomplete answers account for the remaining 4 (2.3\%): the response leaves one requested case or quantity unanswered.

%% file: main/6-conclusion.tex
\section{Conclusion}
We presented \ours, a benchmark of 1,034 PhD-qualifying-exam physics problems with 2,559 reference answers across six domains, each answer judged separately. Our evaluation of 53 models shows that leading open-weight models approach proprietary performance, while most open-weight models solve fewer than half of the test problems without figures. Thinking models occupy the 17 highest ranks. Open-weight multimodal models score 7.3 percentage points lower on average on test problems with figures than on those without. The Qwen3.5-9B error analysis identifies failures in physical modeling and figure interpretation, including figure misreadings that affect \mbox{several subquestions}.

\section*{Acknowledgments}
This work is supported in part by the NVIDIA Academic Grant Program. We are grateful to the Google TRC program for providing computing resources and Together AI for granting \mbox{LLM API credits}.

\section*{Limitations}
\ours evaluates final answers to PhD-qualifying-exam problems across six core physics domains. Future work could broaden coverage to research-level physics and develop criteria for assessing the validity of derivations and proofs. Since the problems come from public sources, their overlap with model pretraining data remains unmeasured. Measuring this overlap and adding newly authored problems would help assess models' ability to solve unfamiliar problems.

%% file: main/Appendix.tex
\appendix

\section{Benchmark Construction and Verification}

\subsection{Author Contributions}
\label{sec:contributions}

\begin{itemize}[topsep=0pt,itemsep=0pt]
\item Project leads: Kaiyue Feng and Yilun Zhao.
\item Initial project idea: Kaiyue Feng, Yilun Zhao, John Sous, and Arman Cohan.
\item Data annotation task management and validation: Kaiyue Feng, Yilun Zhao, and Tianyu Yang. 
\item Experiments: Kaiyue Feng, Yilun Zhao, and Yixin Liu. 
\item Writing: Kaiyue Feng and Yilun Zhao led the writing, with editing contributions from all authors. 
\item Second version revision and dataset verification: Yilun Zhao.
\item Mentorship and advising: Chen Zhao, John Sous, and Arman Cohan.
\end{itemize}

\subsection{Annotator Backgrounds}
\label{Annotation Bib}
\Cref{tab:annotators} summarizes the annotators' qualifications and assigned domains.
\input{figures_tex/annotation_bib}

\input{main/appendix_verification}

\section{Evaluation Protocol and Additional Analyses}

\input{main/appendix_prompts}

\input{main/appendix_results}
\clearpage
\input{main/appendix_errors}

%% file: figures_tex/annotation_bib.tex
\begin{table*}[h]
\centering
\small
\renewcommand{\arraystretch}{1.2}
\setlength{\tabcolsep}{4pt}
\begin{tabular}{@{} c c c >{\raggedright\arraybackslash}p{6cm} @{}}
\toprule
\textbf{ID} & \textbf{Year} & \textbf{Major} & \textbf{Assigned Subject(s)} \\ 
\midrule
1  & 2nd Year Undergraduate & Physics & Classical Mechanics, Electromagnetism \\
2  & 3rd Year Undergraduate & Physics & Quantum Mechanics, Optics \\
3  & 2nd Year Undergraduate & Theoretical Physics & Quantum Mechanics, Thermodynamics and Statistical Mechanics \\
4  & 3rd Year Undergraduate & Applied Physics & Thermodynamics and Statistical Mechanics, Atomic Physics \\
5  & 2nd Year Undergraduate & Engineering Physics & Electromagnetism, Classical Mechanics \\
6  & 3rd Year Undergraduate & Physics & Thermodynamics and Statistical Mechanics, Atomic Physics \\
7  & 2nd Year Undergraduate & Astrophysics & Classical Mechanics, Optics \\
\bottomrule
\end{tabular}
\caption{Biographies of 7 annotators involved in the \ours benchmark construction.}
\label{tab:annotators}
\end{table*}

%% file: main/appendix_verification.tex
\subsection{Verification Details}
\label{sec:verification-details}

\subsubsection{Audit Rules}

Each original problem, with all of its subquestions, formed one audit unit. Only the supplied question text and images were used; agents were instructed not to consult external sources.

\paragraph{Scope.} Subquestions asking for an independently assessable final result were retained. Proof, explanation, and drawing requests were deleted based on the actual requested deliverable, not on keywords. A calculation's supporting derivation was allowed but not counted as a separate proof task. When a retained result depended on a deleted component, the dependency was recorded as either a derivable intermediate step or a missing premise; a missing premise removed the whole problem. If a question depended on a figure that was missing or did not match the problem, the entire problem was removed.

\paragraph{Question text.} Retained question text was kept verbatim. Deleted subquestions were first marked and then omitted from the released questions, with punctuation and whitespace joined at the cut.

\paragraph{Reference answers.} The original reference answers served as the baseline. Corrections required an error established by independent derivation and confirmed by a second verifier. For interpretation, convention, or approximation disputes, the original reference was kept.

\paragraph{Staged process.} A solver started in a fresh context, received only the question and images, and solved the retained subquestions. The workflow used tool-level reference-access rules in a shared filesystem and logged input access. The attempt was frozen before the reference was released. A separate auditor then compared the frozen attempt with the reference. A fresh reviewer checked each batch of up to five problems, including unchanged and removed ones. Rejected items went through another repair and review.

\paragraph{Repairs.} The auditor produced corrected solutions and final answers. Each problem ended as unchanged, repaired, or removed. Repairs followed the original dataset text style and corrected wrong or missing content. Final answers remained concise expressions.

\subsubsection{Verification Funnel}

The first answer-annotation pass covered 1,126 problems. Further scope, dependency, duplicate, and reference checks left 1,109 problems with 2,803 answers. These exclusions, together with earlier filtering, removed 188 of the 1,297 source problems. A final review of answers missed by all five leading models removed 75 further problems (244 answers), producing the released dataset of 1,034 problems with 2,559 answers.

\paragraph{Scope and record checks (188 removals).}
These removals combined several activities: results-only scope filtering removed 93 problems that had no subquestion left to grade. Loss of a required definition or context after subquestion deletion accounted for 79 problems. Missing figures removed 9 problems. The remaining 7 removals were for individual reasons: a corrupted question, a duplicate, a malformed source record, a reference error that invalidated dependent parts, an answer-bearing figure, an inconsistent Hamiltonian, and an unresolved estimate.

\paragraph{Final review (75 removals).}
By category:
\begin{itemize}[nosep]
\item 27 with an incorrect reference answer;
\item 27 with an ambiguous question or answer;
\item 11 with grading ambiguity: 6 involving rounding or unstated approximations, and 5 involving equivalent answers that differed in convention, wording, or representation;
\item 9 with a question defect;
\item 1 with a labeling or format issue.
\end{itemize}

\subsubsection{Repairs}

Among the 1,034 released problems, the main audit corrected reference answers in 245 and worked solutions only in 264 more. An additional 32 problems had formatting-only edits and 20 had clarification-only edits to answers or solutions (e.g., restated conventions); neither category changed a physical result. The answer-changed and solution-only categories are mutually exclusive at the problem level: a problem with both a changed answer and a changed solution is counted under corrected reference answer.

\paragraph{Example.}
Problem \texttt{electro/1\_33} concerns dielectric breakdown voltages in a coaxial cylindrical capacitor. One reference answer gave the maximum voltage as $9.2\times10^3$~V. Re-evaluating $R_2 E_b / (2\sqrt{e})$ gives $9.1\times10^3$~V, so the reference contained an arithmetic error. The answer was updated to $9.1\times10^3$~V and the last line of the worked solution was corrected to match; the other three answers for this problem were kept verbatim.

\subsubsection{Answer Annotation}

Each of the 2,559 retained answers was labeled with its subquestion marker and typed as numeric, symbolic, or text. Two independent annotation passes produced identical labels and types for 1,064 of the 1,126 problems present at that stage (94.5\%), before the later removals. An adjudicator decided the remaining 62.

\subsubsection{Model-Disagreement Screen}

After the main audit, we screened the reference answers that Qwen3.5-9B and Qwen3.8-27B were both judged wrong on (301). On 168 of them, in 129 problems, the two models gave the same answer; these 168 were the leads. Of these, 122 problems (157 answers) were checked by independent derivation, producing 8 proposed errors. A second verifier confirmed 6 of the 8 proposed errors and rejected 2. Five corrections were applied. The sixth problem was removed whole, because its other parts relied on the same wrong result. Three further corrections came from a later audit of judge verdicts, and one bundled answer was split into separate entries. These follow-up checks corrected or split reference answers in 9 problems, of which 7 are in the released dataset.

%% file: main/appendix_prompts.tex
\subsection{Prompts}
\label{sec:prompts}

The response format prompt is shown in \S\ref{Experiment Setup}. It is sent as the system message; for models whose chat template does not accept a system turn, the same text is placed at the start of the user message. The question follows as the user message, with figures attached as images for multimodal models. The judge prompt below is sent once per judged problem to GLM-5.3-Flash; the template variables \texttt{\{question\}}, \texttt{\{response\}}, and \texttt{\{references\}} are filled at runtime.

\begin{tcolorbox}[colback=gray!10,colframe=black!75,title={\small Judge Prompt},breakable]
\footnotesize
\setlength{\baselineskip}{1.2\baselineskip}
Judge whether the following [response] to [question] correctly gives each result in the list of [reference\_answers] below. The reference answers are precise and authoritative.\\[4pt]
{[question]}: \{question\}\\[2pt]
{[response]}: \{response\}\\[2pt]
{[reference\_answers]}:\\
\{references\}\\[4pt]
For EACH reference answer, in order, decide independently:\\[4pt]
\textbf{extracted\_final\_answer}: the response's final answer for the same requested result, copied from its Answers section or, if absent there, from its final statement of that result. When the reference label is a part marker such as (a) or (b)(ii), take only what the response gives for that same part; an answer the response places under a different part does not count. When the label is an ordinal (1, 2, \ldots), the question has no part markers, so identify the answer by the quantity it gives. If the response's answer line for that part states only a choice, a comparison, or a qualitative statement, but its explanation for that same part states the requested quantity, extract that quantity. Use \texttt{None} if the response gives no final answer for it.\\[4pt]
\textbf{reasoning}: explain why the extracted answer matches the reference answer or not, focusing only on meaningful differences. Do not comment on background, do not solve the problem, and do not argue for any answer other than the reference answer.\\[4pt]
\textbf{correct}: \texttt{yes} only if the extracted answer is equivalent to the reference answer under these rules, otherwise \texttt{no}:\\
1.\ Symbolic answers: algebraically equivalent expressions are correct, including rearrangements and notation defined in the question. A missing or different factor, sign, exponent, or variable dependence is incorrect. Quantum state vectors or wavefunctions that differ only by an overall phase factor are equivalent.\\
2.\ Numeric answers: correct if within 1\% relative difference of the reference value, or equal to it when rounded to the reference's significant figures. For order-of-magnitude estimates, the same order of magnitude is correct. Units must be equivalent (unit conversions are fine); a missing unit is incorrect unless the question fixes the unit.\\
3.\ If the reference states a formula followed by its value, compare the final result; the response must not contradict either.\\
4.\ Verbal answers: correct if they state the same conclusion, choice, or identification. Extra correct detail is fine; any contradiction is incorrect. When the reference gives a formula, value, or explicit result, a qualitative description of it is incorrect unless the response also states that formula or value for the same part.\\
5.\ Incorrect if the response gives several different candidate answers for one result, hedges between alternatives, answers a different quantity, gives the result only under a different part, or gives no final answer. Intermediate steps shown in the reference are not required.\\
6.\ An answer that equals the reference once a convention or conversion the response itself states is applied is correct. The response must state the convention, and the two must then agree exactly; a convention it does not state, or one that still leaves a difference in factor, sign, exponent, or variable dependence, does not make a wrong answer correct.\\[4pt]
Also extract confidence: the confidence score between 0 and 100 stated in the response; use 100 if none is given.\\*[4pt]
Return only json in exactly this shape, with one entry per reference answer in the same order:\\*
\texttt{\{"answers": [\{"index": 0, "label": "(a)", "extracted\_final\_answer": "...", "reasoning": "...", "correct": "yes"\}], "confidence": 85\}}
\end{tcolorbox}

\subsection{Judge Validation}
\label{sec:judge-validation}

\paragraph{Synthetic self-test.}
The self-test builds synthetic responses from the reference answers, so every verdict is known in advance. In each response one answer is the target. It is either left as the reference, shifted by 0.5\% (still correct under the 1\% tolerance), or made wrong: a doubled value, a formula multiplied by 2, a hedge between the right and a wrong answer, a missing answer, or two labeled parts swapped. Non-target answers are copied unchanged, except the second answer in a swapped-part case. The two misses reported in \S\ref{Automated Evaluation System} both accepted a hedged text answer.

\paragraph{Cross-check with DeepSeek-V4.1-Flash.}
The sample is a random draw of 200 problems answered by Qwen3.5-9B. The four human reviewers saw each disagreement with the judges labeled only A and B. In the remaining 2 disagreements, DeepSeek-V4.1-Flash accepted an answer that GLM-5.3-Flash rejected.

\paragraph{Independent grading of GLM-5.3-Flash.}
We also graded GLM-5.3-Flash's responses with DeepSeek-V4.1-Flash using the same evaluation protocol. \Cref{tab:glm-independent} reports its scores and their differences from the GLM-5.3-Flash-judged scores on each population.

\begin{table}[h]
\centering
\small
\begin{tabular}{lrrrrr}
\toprule
Population & Problems & Strict (\%) & CI low & CI high & Gap (pp) \\
\midrule
Val (no fig.) & 184 & 82.1 & 76.6 & 87.0 & 6.5 \\
Val (all) & 240 & 81.2 & 75.8 & 85.8 & 6.2 \\
Test (no fig.) & 602 & 80.6 & 77.4 & 83.7 & 5.0 \\
Test (all) & 794 & 80.5 & 77.8 & 83.4 & 4.8 \\
\bottomrule
\end{tabular}
\caption{GLM-5.3-Flash strict accuracy when judged by DeepSeek-V4.1-Flash. CI: 95\% bootstrap interval over problems. Gap: GLM-5.3-Flash-judged minus DeepSeek-V4.1-Flash-judged strict accuracy on the \mbox{same population, in percentage points}.}
\label{tab:glm-independent}
\end{table}

%% file: main/appendix_results.tex
\subsection{Full Results}
\label{sec:full-results}

\Cref{tab:full-results} reports strict, answer-level, and partial-credit accuracy for all 53 models on the validation and test splits, together with scores on all 794 test problems for the 26 multimodal models.

\input{figures_tex/generated/full_results}

\subsection{Additional Analyses}
\label{sec:additional-analyses}

The comparisons use released checkpoints spanning different parameter counts, architectures, generations, modalities, quantization settings, and thinking status. Unless stated otherwise, intervals are 95\% bootstrap intervals from resampling problems. Domain-gap intervals are pointwise and conditional on the model pair selected by overall figure-free test accuracy. Percentages and score gaps are rounded to one decimal after computation.

\subsubsection{Model Scale}
\label{sec:appendix-scale}

\begin{figure}[t]
  \centering
  \includegraphics[width=0.6\linewidth]{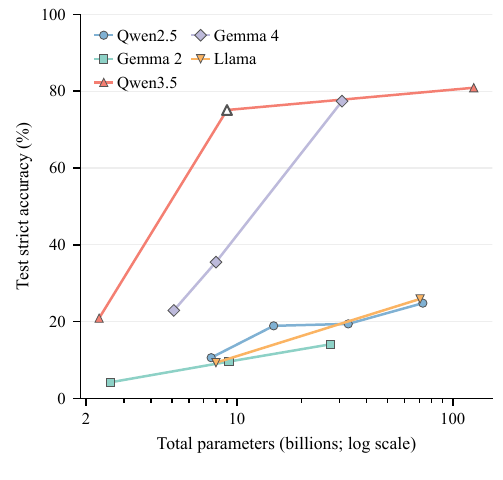}
  \caption{Figure-free test strict accuracy versus total parameters (log scale) within five model families. The hollow marker for Qwen3.5-9B uses its nominal parameter count. Gemma E models use total parameters. Qwen3.5's largest model is MoE; the Llama pair changes generation.}
  \label{fig:analysis-scale}
\end{figure}

\Cref{fig:analysis-scale} shows strict accuracy against total parameter count within five model families. Accuracy rises between the endpoints in each family: Qwen3.5 gains 60.0~pp [56.0, 64.1] from 2B to 122B, Gemma 4 gains 54.5~pp [50.0, 58.8] from E2B to 31B, and the older families (Qwen2.5, Gemma 2, Llama) gain 10.0--16.6~pp. Most of the Qwen3.5 gain comes from the 2B-to-9B step (54.2 of 60.0~pp), while Gemma 4 gains more in its final step to 31B (41.9~pp of 54.5~pp). The Qwen2.5-14B-Instruct to Qwen2.5-32B-Instruct step changes accuracy by 0.5~pp [$-$2.3, 3.3].

\subsubsection{Successive Checkpoints from the Same Developer}
\label{sec:appendix-generations}

\Cref{tab:analysis_replacements} pairs each older checkpoint with a newer checkpoint from the same developer. A newer checkpoint can appear in several pairs, linking it to different predecessors.

\input{figures_tex/generated/analysis_generations_table}

\subsubsection{Completion Length}
\label{sec:appendix-length}

\begin{figure}[htbp]
  \centering
  \includegraphics[width=0.6\linewidth]{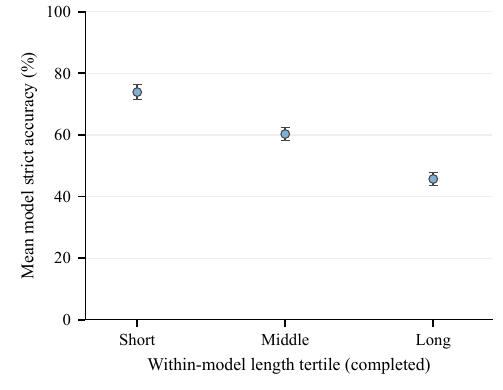}
  \caption{Mean strict accuracy within short, middle, and long completed-response tertiles, averaged over 21 thinking models, with problem-resampling 95\% intervals.}
  \label{fig:analysis-length}
\end{figure}

\Cref{fig:analysis-length} examines the association between completed-response length and accuracy. For each of the 21 thinking models, responses that finished before the output limit are sorted by reported completion tokens and split into equally sized rank tertiles. Strict accuracy is computed in each tertile and then averaged equally over models. The shortest-response tertile averages 73.9\% [71.4, 76.5], the middle 60.3\% [58.3, 62.5], and the longest 45.7\% [43.6, 47.8]. Longer completed responses are associated with lower accuracy.

\subsubsection{Domain Analysis}
\label{sec:appendix-domains}

\begin{figure}[t]
  \centering
  \subfigure[Mean strict accuracy by domain on figure-free test problems, averaged over all 53 models, with problem-resampling 95\% intervals.]{\includegraphics[width=0.48\linewidth]{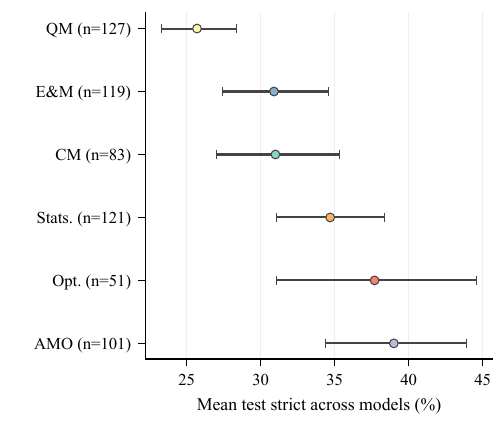}\label{fig:analysis-domains}}\hfill
  \subfigure[Overall and domain-level strict accuracy gap between GPT-6 Astra and Qwen3.8-27B on figure-free test problems, with problem-resampling 95\% intervals. The overall estimate is dark gray; domain estimates use the domain colors.]{\includegraphics[width=0.48\linewidth]{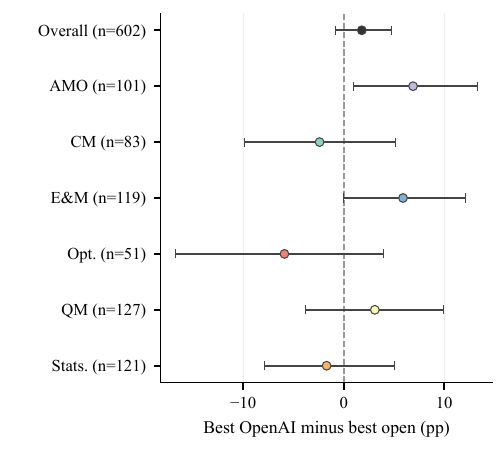}\label{fig:analysis-ownership}}
  \caption{Domain-level analyses on figure-free test problems.}
  \label{fig:domain-analyses}
\end{figure}

\Cref{fig:analysis-domains} shows the mean strict accuracy by domain, averaged over the 53 models. Quantum Mechanics is the lowest at 25.7\% [23.3, 28.4], followed by Electromagnetism at 30.9\% [27.4, 34.6] and Classical Mechanics at 31.0\% [27.0, 35.3]. At the answer level, Electromagnetism (38.8\%) and Classical Mechanics (39.3\%) are lowest and Quantum Mechanics is third lowest (43.6\%). The strict ranking partly reflects answer count: Quantum Mechanics problems carry 2.91 answers on average (the highest), and strict scoring requires every answer correct. Among the 10 strongest models by overall figure-free strict, Optics is lowest (80.4\%) and Quantum Mechanics second lowest (84.0\%); the six domains span 80.4--86.3\% for these models. Optics has 51 figure-free test problems and the widest interval.

\subsubsection{Best OpenAI and Best Open-Weight Model by Domain}
\label{sec:appendix-ownership}

The best OpenAI model (GPT-6 Astra, 90.4\%) and the best open-weight model (Qwen3.8-27B, 88.5\%) differ by 1.8~pp [$-$0.8, 4.8] on figure-free test strict. \Cref{fig:analysis-ownership} shows this gap by domain. Qwen3.8-27B scores higher in Classical Mechanics, Thermodynamics and Statistical Mechanics, and Optics, while GPT-6 Astra leads by 6.9~pp [1.0, 13.3] in Atomic Physics and 5.9~pp [0.0, 12.1] in Electromagnetism.

\subsubsection{Truncation Diagnostics}
\label{sec:appendix-truncation}

\begin{figure}[t]
  \centering
  \includegraphics[width=0.6\linewidth]{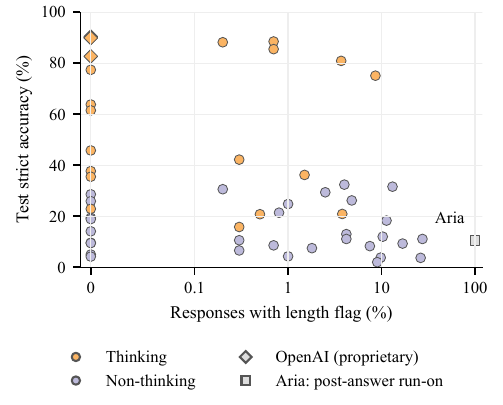}
  \caption{Figure-free test strict accuracy versus share of responses that stopped at the output token limit. The symmetric-log axis has a linear threshold of 0.1\%; color denotes thinking status, diamonds the four OpenAI models. Aria (gray square) has a separate marker for its post-answer run-on and is \mbox{excluded from the non-thinking truncation pool}.}
  \label{fig:analysis-truncation}
\end{figure}

\Cref{fig:analysis-truncation} plots figure-free test strict accuracy against the share of responses that stopped at the output token limit. Aria's length flags reflect post-answer run-on caused by a serving stop-token error, and it is shown separately. Among the 10 lowest-scoring thinking models, 37 of 3,977 failures (0.9\%) are truncated. The six non-thinking models other than Aria with more than 10\% of responses truncated are Ministral-3-8B-Instruct-2512, InternVL2.5-38B, Qwen2.5-Math-7B, Qwen2-VL-72B-Instruct-AWQ, Llama-3.1-8B-Instruct, and DeepSeek-VL2-Small.

%% file: figures_tex/generated/full_results.tex
\begingroup
\small
\setlength{\tabcolsep}{3pt}
\renewcommand{\arraystretch}{0.95}
\setlength{\LTleft}{\fill}
\setlength{\LTright}{\fill}
\begin{longtable}{@{}l c r r r r r r r r r@{}}
\caption{Full results for all 53 models (\%). S: strict accuracy, requiring every answer in a problem to be correct. A: answer-level accuracy, pooling all answers. P: partial credit, averaging the fraction of correct answers within each problem with equal problem weights. Val and Test (no fig.) exclude figure problems; Test (all) includes test problems with and without figures and applies only to multimodal models; $^{\dagger}$ marks text-only models, with -- for Test (all). Think uses \cmark{}/\xmark{} for thinking/other models. Models are ordered within each group by \mbox{descending strict Test (no fig.) accuracy}.}\label{tab:full-results} \\
\toprule
\textbf{Model} & \textbf{Think} & \multicolumn{3}{c}{\textbf{Val (no fig.)}} & \multicolumn{3}{c}{\textbf{Test (no fig.)}} & \multicolumn{3}{c}{\textbf{Test (all)}} \\
 &  & \multicolumn{3}{c}{\footnotesize (184 problems)} & \multicolumn{3}{c}{\footnotesize (602 problems)} & \multicolumn{3}{c}{\footnotesize (794 problems)} \\
\cmidrule(lr){3-5}\cmidrule(lr){6-8}\cmidrule(l){9-11}
 &  & \textbf{S} & \textbf{A} & \textbf{P} & \textbf{S} & \textbf{A} & \textbf{P} & \textbf{S} & \textbf{A} & \textbf{P} \\
\endfirsthead
\multicolumn{11}{c}{\tablename~\thetable{} (continued)} \\
\toprule
\textbf{Model} & \textbf{Think} & \multicolumn{3}{c}{\textbf{Val (no fig.)}} & \multicolumn{3}{c}{\textbf{Test (no fig.)}} & \multicolumn{3}{c}{\textbf{Test (all)}} \\
 &  & \multicolumn{3}{c}{\footnotesize (184 problems)} & \multicolumn{3}{c}{\footnotesize (602 problems)} & \multicolumn{3}{c}{\footnotesize (794 problems)} \\
\cmidrule(lr){3-5}\cmidrule(lr){6-8}\cmidrule(l){9-11}
 &  & \textbf{S} & \textbf{A} & \textbf{P} & \textbf{S} & \textbf{A} & \textbf{P} & \textbf{S} & \textbf{A} & \textbf{P} \\
\endhead
\midrule
\multicolumn{11}{r}{\footnotesize Continued on the next page.} \\
\endfoot
\bottomrule
\endlastfoot
\midrule
\multicolumn{11}{c}{\textbf{Proprietary Models}} \\
GPT-6 Astra\vphantom{\cmark} & \cmark & 92.4 & 96.2 & 96.1 & 90.4 & 94.9 & 94.6 & 91.1 & 95.2 & 95.0 \\
GPT-6 Sol\vphantom{\cmark} & \cmark & 90.2 & 96.0 & 94.9 & 90.0 & 95.1 & 94.2 & 90.6 & 95.1 & 94.6 \\
GPT-5.6 Sol\vphantom{\cmark} & \cmark & 90.8 & 95.8 & 95.0 & 89.9 & 94.9 & 93.5 & 90.4 & 95.0 & 93.8 \\
GPT-6 Luna\vphantom{\cmark} & \cmark & 83.2 & 88.8 & 89.6 & 82.7 & 90.0 & 89.2 & 82.6 & 89.7 & 89.5 \\
\midrule
\multicolumn{11}{c}{\textbf{Open-Weight Models}} \\
Qwen3.8-27B\vphantom{\cmark} & \cmark & 87.5 & 93.2 & 92.5 & 88.5 & 92.7 & 92.6 & 87.2 & 91.7 & 91.8 \\
DeepSeek-V4.1-Flash\vphantom{\cmark} & \cmark & 91.8 & 95.6 & 95.8 & 88.2 & 93.3 & 93.4 & 87.0 & 92.6 & 92.8 \\
GLM-5.3-Flash\vphantom{\cmark} & \cmark & 88.6 & 94.5 & 94.0 & 85.5 & 91.2 & 90.9 & 85.3 & 91.2 & 90.8 \\
Qwen3.5-122B-A10B\vphantom{\cmark} & \cmark & 85.3 & 90.5 & 91.1 & 80.9 & 89.0 & 88.8 & 80.7 & 89.1 & 88.5 \\
gemma-4-31B-it\vphantom{\cmark} & \cmark & 83.2 & 89.0 & 90.1 & 77.4 & 86.9 & 85.7 & 76.8 & 86.2 & 85.2 \\
Qwen3.5-9B\vphantom{\cmark} & \cmark & 78.3 & 85.0 & 84.4 & 75.1 & 83.8 & 82.8 & 73.8 & 82.8 & 82.0 \\
Mistral-Small-4-119B-2603\vphantom{\cmark} & \cmark & 72.8 & 81.4 & 82.6 & 63.8 & 78.1 & 78.1 & 61.7 & 74.8 & 75.5 \\
DeepSeek-R1\smash{$^{\dagger}$}\vphantom{\cmark} & \cmark & 65.2 & 79.2 & 79.1 & 61.5 & 74.9 & 74.1 & -- & -- & -- \\
DeepSeek-R1-0528-Qwen3-8B\smash{$^{\dagger}$}\vphantom{\cmark} & \cmark & 56.0 & 68.4 & 71.0 & 45.8 & 61.8 & 61.7 & -- & -- & -- \\
DeepSeek-R1-Distill-Qwen-32B\smash{$^{\dagger}$}\vphantom{\cmark} & \cmark & 45.1 & 60.2 & 60.5 & 42.2 & 58.0 & 57.6 & -- & -- & -- \\
Intern-S1-mini\vphantom{\cmark} & \cmark & 40.8 & 53.2 & 55.6 & 37.7 & 51.7 & 51.9 & 34.4 & 47.4 & 48.0 \\
QwQ-32B-Preview\smash{$^{\dagger}$}\vphantom{\cmark} & \cmark & 34.2 & 51.5 & 50.1 & 36.2 & 51.4 & 50.2 & -- & -- & -- \\
gemma-4-E4B-it\vphantom{\cmark} & \cmark & 33.7 & 49.4 & 50.6 & 35.5 & 51.0 & 50.9 & 31.7 & 47.3 & 47.0 \\
Ministral-3-14B-Instruct-2512\vphantom{\cmark} & \xmark & 37.0 & 55.1 & 53.7 & 32.4 & 51.0 & 48.2 & 29.6 & 47.1 & 45.2 \\
Ministral-3-8B-Instruct-2512\vphantom{\cmark} & \xmark & 31.0 & 45.1 & 46.1 & 31.6 & 45.7 & 45.6 & 27.7 & 41.7 & 41.7 \\
Llama-4-Scout-17B-16E-Instruct\vphantom{\cmark} & \xmark & 35.9 & 48.7 & 52.1 & 30.6 & 45.9 & 45.3 & 28.2 & 43.3 & 42.8 \\
Qwen2.5-Math-72B-Instruct\smash{$^{\dagger}$}\vphantom{\cmark} & \xmark & 34.8 & 47.7 & 50.1 & 29.4 & 44.1 & 43.3 & -- & -- & -- \\
Phi-4\smash{$^{\dagger}$}\vphantom{\cmark} & \xmark & 33.2 & 46.0 & 47.1 & 28.6 & 43.5 & 43.4 & -- & -- & -- \\
InternVL3.5-38B\vphantom{\cmark} & \xmark & 29.9 & 44.9 & 44.7 & 26.2 & 42.1 & 41.3 & 24.6 & 39.4 & 39.2 \\
Llama-3.3-70B-Instruct\smash{$^{\dagger}$}\vphantom{\cmark} & \xmark & 29.9 & 43.6 & 44.1 & 25.9 & 41.7 & 40.3 & -- & -- & -- \\
Qwen2.5-72B-Instruct\smash{$^{\dagger}$}\vphantom{\cmark} & \xmark & 27.2 & 43.2 & 42.6 & 24.8 & 40.0 & 38.6 & -- & -- & -- \\
gemma-4-E2B-it\vphantom{\cmark} & \cmark & 19.6 & 35.6 & 34.8 & 22.9 & 39.8 & 38.3 & 20.2 & 35.4 & 34.5 \\
Mistral-Small-24B-Instruct-2501\smash{$^{\dagger}$}\vphantom{\cmark} & \xmark & 21.2 & 33.7 & 35.1 & 21.4 & 36.7 & 35.2 & -- & -- & -- \\
Qwen3.5-2B\vphantom{\cmark} & \cmark & 19.0 & 34.5 & 33.8 & 20.9 & 33.6 & 33.1 & 19.4 & 31.4 & 31.0 \\
Ministral-3-8B-Reasoning-2512\vphantom{\cmark} & \cmark & 20.7 & 37.1 & 35.7 & 20.8 & 36.7 & 34.0 & 18.8 & 33.2 & 31.3 \\
Qwen2.5-32B-Instruct\smash{$^{\dagger}$}\vphantom{\cmark} & \xmark & 22.3 & 36.7 & 35.4 & 19.4 & 34.3 & 32.9 & -- & -- & -- \\
Qwen2.5-14B-Instruct\smash{$^{\dagger}$}\vphantom{\cmark} & \xmark & 17.4 & 31.1 & 30.9 & 18.9 & 30.7 & 30.9 & -- & -- & -- \\
InternVL2.5-38B\vphantom{\cmark} & \xmark & 19.0 & 32.0 & 32.3 & 18.3 & 33.8 & 31.5 & 16.4 & 30.6 & 28.8 \\
Phi-4-Reasoning-Vision-15B\vphantom{\cmark} & \cmark & 17.9 & 30.9 & 31.7 & 15.8 & 30.7 & 28.4 & 14.6 & 27.3 & 26.2 \\
gemma-2-27b-it\smash{$^{\dagger}$}\vphantom{\cmark} & \xmark & 10.3 & 21.0 & 19.2 & 14.1 & 27.7 & 25.1 & -- & -- & -- \\
Qwen2.5-Math-7B-Instruct\smash{$^{\dagger}$}\vphantom{\cmark} & \xmark & 15.8 & 23.9 & 26.5 & 13.0 & 25.4 & 24.1 & -- & -- & -- \\
Qwen2.5-Math-7B\smash{$^{\dagger}$}\vphantom{\cmark} & \xmark & 10.3 & 22.2 & 20.9 & 12.0 & 23.0 & 20.8 & -- & -- & -- \\
Qwen2-VL-72B-Instruct-AWQ\vphantom{\cmark} & \xmark & 11.4 & 20.1 & 21.2 & 11.1 & 21.3 & 20.3 & 9.7 & 19.2 & 18.2 \\
DeepSeekMath-7B-RL\smash{$^{\dagger}$}\vphantom{\cmark} & \xmark & 6.0 & 16.7 & 15.7 & 11.1 & 22.0 & 20.3 & -- & -- & -- \\
Qwen2.5-7B-Instruct\smash{$^{\dagger}$}\vphantom{\cmark} & \xmark & 7.1 & 19.7 & 17.3 & 10.6 & 22.3 & 20.3 & -- & -- & -- \\
Aria\vphantom{\cmark} & \xmark & 5.4 & 18.0 & 15.1 & 10.5 & 21.7 & 19.5 & 9.2 & 18.8 & 17.2 \\
gemma-2-9b-it\smash{$^{\dagger}$}\vphantom{\cmark} & \xmark & 6.0 & 17.4 & 16.2 & 9.6 & 21.5 & 19.3 & -- & -- & -- \\
Yi-1.5-34B-Chat\smash{$^{\dagger}$}\vphantom{\cmark} & \xmark & 8.7 & 19.1 & 19.3 & 9.5 & 20.8 & 18.7 & -- & -- & -- \\
Llama-3.1-8B-Instruct\smash{$^{\dagger}$}\vphantom{\cmark} & \xmark & 7.6 & 16.7 & 17.1 & 9.3 & 19.9 & 18.4 & -- & -- & -- \\
InternLM3-8B-Instruct-AWQ\smash{$^{\dagger}$}\vphantom{\cmark} & \xmark & 11.4 & 19.5 & 20.1 & 8.6 & 20.8 & 18.1 & -- & -- & -- \\
Qwen2.5-Math-1.5B-Instruct\smash{$^{\dagger}$}\vphantom{\cmark} & \xmark & 6.0 & 14.2 & 13.2 & 8.3 & 16.0 & 15.7 & -- & -- & -- \\
Mathstral-7B-v0.1\smash{$^{\dagger}$}\vphantom{\cmark} & \xmark & 3.3 & 18.2 & 14.3 & 7.5 & 18.7 & 15.5 & -- & -- & -- \\
Pixtral-12B-2409\vphantom{\cmark} & \xmark & 8.2 & 16.7 & 15.9 & 6.6 & 18.0 & 15.0 & 5.4 & 16.1 & 13.4 \\
c4ai-command-r-v01\smash{$^{\dagger}$}\vphantom{\cmark} & \xmark & 2.7 & 10.6 & 9.7 & 5.0 & 12.7 & 10.8 & -- & -- & -- \\
Mistral-7B-Instruct-v0.3\smash{$^{\dagger}$}\vphantom{\cmark} & \xmark & 4.9 & 11.4 & 10.7 & 4.3 & 13.7 & 11.0 & -- & -- & -- \\
gemma-2-2b-it\smash{$^{\dagger}$}\vphantom{\cmark} & \xmark & 3.3 & 6.6 & 6.8 & 4.2 & 8.5 & 7.8 & -- & -- & -- \\
GLM-4-9B-Chat-hf\smash{$^{\dagger}$}\vphantom{\cmark} & \xmark & 4.3 & 9.7 & 9.5 & 3.8 & 11.0 & 9.3 & -- & -- & -- \\
DeepSeek-VL2-Small\vphantom{\cmark} & \xmark & 2.2 & 6.1 & 6.4 & 3.7 & 7.8 & 7.1 & 3.0 & 7.0 & 6.3 \\
ChatGLM3-6B\smash{$^{\dagger}$}\vphantom{\cmark} & \xmark & 3.3 & 5.5 & 6.1 & 2.0 & 3.4 & 3.6 & -- & -- & -- \\
\end{longtable}
\endgroup

%% file: figures_tex/generated/analysis_generations_table.tex
\begingroup
\small
\setlength{\tabcolsep}{4pt}
\begin{longtable}{@{}>{\raggedright\arraybackslash}p{0.27\textwidth}>{\raggedright\arraybackslash}p{0.27\textwidth}rrr@{}}
\caption{Successive checkpoints from the same developer on figure-free test problems. Pairs share successors and often change size, modality or thinking. Intervals resample problems with checkpoints fixed.}\label{tab:analysis_replacements}\\
\toprule
Older & Mapped successor & Older (\%) & New (\%) & Gain (pp) [95\% CI] \\
\midrule
\endfirsthead
\toprule
Older & Mapped successor & Older (\%) & New (\%) & Gain (pp) [95\% CI] \\
\midrule
\endhead
\midrule
\multicolumn{5}{r}{Continued on next page}\\
\endfoot
\bottomrule
\endlastfoot
Qwen2.5-\allowbreak{}Math-\allowbreak{}1.5B-\allowbreak{}Instruct & Qwen3.5-\allowbreak{}2B & 8.3 & 20.9 & +12.6 [9.5, 15.9] \\
gemma-\allowbreak{}2-\allowbreak{}2b-\allowbreak{}it & gemma-\allowbreak{}4-\allowbreak{}E2B-\allowbreak{}it & 4.2 & 22.9 & +18.8 [15.3, 22.3] \\
gemma-\allowbreak{}2-\allowbreak{}9b-\allowbreak{}it & gemma-\allowbreak{}4-\allowbreak{}E4B-\allowbreak{}it & 9.6 & 35.5 & +25.9 [22.3, 29.9] \\
InternLM3-\allowbreak{}8B-\allowbreak{}Instruct-\allowbreak{}AWQ & Intern-\allowbreak{}S1-\allowbreak{}mini & 8.6 & 37.7 & +29.1 [25.4, 33.2] \\
Phi-\allowbreak{}4 & Phi-\allowbreak{}4-\allowbreak{}Reasoning-\allowbreak{}Vision-\allowbreak{}15B & 28.6 & 15.8 & $-$12.8 [$-$16.1, $-$9.6] \\
DeepSeek-\allowbreak{}R1-\allowbreak{}Distill-\allowbreak{}Qwen-\allowbreak{}32B & DeepSeek-\allowbreak{}R1-\allowbreak{}0528-\allowbreak{}Qwen3-\allowbreak{}8B & 42.2 & 45.8 & +3.7 [$-$0.3, 7.6] \\
DeepSeekMath-\allowbreak{}7B-\allowbreak{}RL & DeepSeek-\allowbreak{}R1-\allowbreak{}0528-\allowbreak{}Qwen3-\allowbreak{}8B & 11.1 & 45.8 & +34.7 [30.4, 39.2] \\
InternVL2.5-\allowbreak{}38B & InternVL3.5-\allowbreak{}38B & 18.3 & 26.2 & +8.0 [4.8, 11.1] \\
Mistral-\allowbreak{}7B-\allowbreak{}Instruct-\allowbreak{}v0.3 & Ministral-\allowbreak{}3-\allowbreak{}8B-\allowbreak{}Instruct-\allowbreak{}2512 & 4.3 & 31.6 & +27.2 [23.8, 30.9] \\
Mathstral-\allowbreak{}7B-\allowbreak{}v0.1 & Ministral-\allowbreak{}3-\allowbreak{}8B-\allowbreak{}Reasoning-\allowbreak{}2512 & 7.5 & 20.8 & +13.3 [10.1, 16.6] \\
Pixtral-\allowbreak{}12B-\allowbreak{}2409 & Ministral-\allowbreak{}3-\allowbreak{}14B-\allowbreak{}Instruct-\allowbreak{}2512 & 6.6 & 32.4 & +25.7 [22.3, 29.6] \\
Mistral-\allowbreak{}Small-\allowbreak{}24B-\allowbreak{}Instruct-\allowbreak{}2501 & Mistral-\allowbreak{}Small-\allowbreak{}4-\allowbreak{}119B-\allowbreak{}2603 & 21.4 & 63.8 & +42.4 [37.9, 46.7] \\
Llama-\allowbreak{}3.3-\allowbreak{}70B-\allowbreak{}Instruct & Llama-\allowbreak{}4-\allowbreak{}Scout-\allowbreak{}17B-\allowbreak{}16E-\allowbreak{}Instruct & 25.9 & 30.6 & +4.7 [1.3, 8.1] \\
Qwen2.5-\allowbreak{}72B-\allowbreak{}Instruct & Qwen3.5-\allowbreak{}122B-\allowbreak{}A10B & 24.8 & 80.9 & +56.1 [52.0, 60.3] \\
Qwen2.5-\allowbreak{}Math-\allowbreak{}72B-\allowbreak{}Instruct & Qwen3.5-\allowbreak{}122B-\allowbreak{}A10B & 29.4 & 80.9 & +51.5 [47.3, 55.8] \\
Qwen2-\allowbreak{}VL-\allowbreak{}72B-\allowbreak{}Instruct-\allowbreak{}AWQ & Qwen3.5-\allowbreak{}122B-\allowbreak{}A10B & 11.1 & 80.9 & +69.8 [65.9, 73.6] \\
Qwen2.5-\allowbreak{}7B-\allowbreak{}Instruct & Qwen3.5-\allowbreak{}9B & 10.6 & 75.1 & +64.5 [60.5, 68.4] \\
Qwen2.5-\allowbreak{}14B-\allowbreak{}Instruct & Qwen3.5-\allowbreak{}9B & 18.9 & 75.1 & +56.1 [51.8, 60.5] \\
Qwen2.5-\allowbreak{}Math-\allowbreak{}7B & Qwen3.5-\allowbreak{}9B & 12.0 & 75.1 & +63.1 [59.1, 67.3] \\
Qwen2.5-\allowbreak{}32B-\allowbreak{}Instruct & Qwen3.8-\allowbreak{}27B & 19.4 & 88.5 & +69.1 [65.4, 72.8] \\
QwQ-\allowbreak{}32B-\allowbreak{}Preview & Qwen3.8-\allowbreak{}27B & 36.2 & 88.5 & +52.3 [48.3, 56.5] \\
gemma-\allowbreak{}2-\allowbreak{}27b-\allowbreak{}it & gemma-\allowbreak{}4-\allowbreak{}31B-\allowbreak{}it & 14.1 & 77.4 & +63.3 [59.1, 67.1] \\
GLM-\allowbreak{}4-\allowbreak{}9B-\allowbreak{}Chat-\allowbreak{}hf & GLM-\allowbreak{}5.3-\allowbreak{}Flash & 3.8 & 85.5 & +81.7 [78.7, 84.9] \\
ChatGLM3-\allowbreak{}6B & GLM-\allowbreak{}5.3-\allowbreak{}Flash & 2.0 & 85.5 & +83.6 [80.7, 86.5] \\
DeepSeek-\allowbreak{}R1 & DeepSeek-\allowbreak{}V4.1-\allowbreak{}Flash & 61.5 & 88.2 & +26.7 [23.1, 30.6] \\
DeepSeek-\allowbreak{}VL2-\allowbreak{}Small & DeepSeek-\allowbreak{}V4.1-\allowbreak{}Flash & 3.7 & 88.2 & +84.6 [81.9, 87.2] \\
\end{longtable}
\endgroup

%% file: main/appendix_errors.tex
\raggedbottom
\section{Error Analysis Examples}
\label{sec:appendix-errors}

This appendix shows one Qwen3.5-9B error for each of the six main categories in \Cref{sec:error-analysis}.

\subsection{Lack of Professional Knowledge}
\label{appendix:knowledge}
\input{figures_tex/Professional_knowledge}

\subsection{Figure Misreading}
\label{appendix:figure}
\input{figures_tex/vision_error}

\subsection{Calculation Errors}
\label{appendix:calculation}
\input{figures_tex/calculation_error}

\subsection{Answer-Form Mismatch}
\label{appendix:answer-form}

\begin{tcolorbox}[colframe=black,colback=lightgrey]
\textbf{Question:}

[\ldots] Suppose a 200~GeV/$c$ $\pi$ meson beam is used to produce neutrinos via the decay $\pi \rightarrow \mu + \nu$. [\ldots] (c)~Calculate the minimum and maximum momenta the neutrinos can have.\\

\textbf{Reference answer:}

$p_\text{min} = 1.04 \times 10^{-2}$~MeV/$c$;\quad $p_\text{max} = 8.54 \times 10^{4}$~MeV/$c$.
\end{tcolorbox}

\begin{tcolorbox}[colframe=black,colback=white,title=Qwen3.5-9B]
``[\ldots] $p_{\nu,\text{min}} = \gamma_\pi p_\nu^* (1 - \beta_\pi) \approx p_\nu^* / (2\gamma_\pi) \approx \frac{29.78}{2865} \approx 0.0104$~MeV/$c$ $\approx 10$~keV/$c$.
Given the context, the minimum is effectively 0, but the kinematic limit is $\approx 10$~keV/$c$. I will report the value close to 0.''

Final answer: ``85.3~GeV/$c$ and $\approx 0$~GeV/$c$ (or 10~keV/$c$).''

\begin{tcolorbox}[colback=red!10, colframe=red!80, boxrule=0.75pt]
\textbf{Error:} The model correctly derives $p_\text{min} \approx 10$~keV/$c$ in its working but then reports ``$\approx 0$'' in the final answer, hedging between two conflicting values. The grading rules reject hedged or conflicting final answers.
\end{tcolorbox}
\end{tcolorbox}

\begin{tcolorbox}[colframe=black,colback=lightred,title=Error Category: Answer-Form Mismatch]
\textbf{Error reason:} The physics is correct, but the final answer hedges between two conflicting values for the minimum momentum.
\end{tcolorbox}

\subsection{Non-Existent Assumptions}
\label{appendix:assumption}
\input{figures_tex/assumption}

\subsection{Misunderstanding the Question}
\label{appendix:misunderstanding}
\input{figures_tex/misunderstanding}

\flushbottom

%% file: figures_tex/Professional_knowledge.tex
\begin{tcolorbox}[colframe=black,colback=lightgrey]
\textbf{Question:}

An iron rod of length $L$ and circular cross section (radius $R$), is given a uniform longitudinal magnetization $\mathbf{M}$, and then bent around into a circle with a narrow gap (width $w$). Find the magnetic field at the center of the gap, assuming $w \ll R \ll L$.\\

\textbf{Reference answer:}\smallskip

$\mathbf{B} = \mu_0 M \left(1 - \frac{w}{2R}\right)$.
\end{tcolorbox}

\begin{tcolorbox}[colframe=black,colback=white,title=Qwen3.5-9B]
``At the gap, the cut ends of the rod act as two parallel sheets of current with surface current density $K = M$. One sheet carries current in one vertical direction, and the other in the opposite vertical direction. The magnetic field between two such parallel current sheets with opposite currents~$K$ is the sum of the fields from each sheet, which is $\mu_0 K$. Since $K=M$, the field magnitude is $\mu_0 M$. The direction is opposite to $\mathbf{M}$.''

\begin{tcolorbox}[colback=red!10, colframe=red!80, boxrule=0.75pt]
\textbf{Error:} The magnetization is longitudinal (parallel to the rod axis), so it is normal to the cut faces. The bound surface current $\mathbf{M}\times\hat{\mathbf{n}}$ on those faces is therefore zero. Inventing opposing current sheets there reverses the predicted field direction and omits the $-w/2R$ correction from the missing gap.
\end{tcolorbox}
\end{tcolorbox}

\begin{tcolorbox}[colframe=black,colback=lightred,title=Error Category: Lack of Professional Knowledge]
\textbf{Error reason:} The model applies the wrong physical model for bound surface currents, producing a field with the incorrect sign.
\end{tcolorbox}

%% file: figures_tex/vision_error.tex
\begin{figure}[H]
    \centering
    \includegraphics[width=0.18\linewidth]{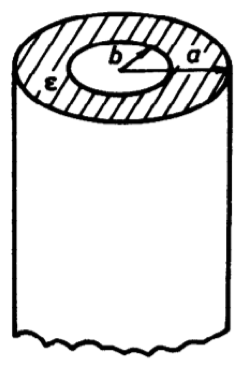}
    \caption{The figure given with the question. The inner radius is labeled~$b$ and the outer radius~$a$.}
    \label{fig:error-figure-cylinders}
\end{figure}

\begin{tcolorbox}[colframe=black,colback=lightgrey]
\textbf{Question:}

The space between two long thin metal cylinders is filled with a material with dielectric constant $\epsilon$. The cylinders have radii $a$ and $b$, as shown in \Cref{fig:error-figure-cylinders}.

(a) What is the charge per unit length on the cylinders when the potential between them is $V$ with the outer cylinder at the higher potential?

(b) What is the electric field between the cylinders?\\

\textbf{Reference answer:}

The capacitance per unit length is $C = 2\pi\epsilon / \ln(a/b)$, giving charges $\lambda_i = -2\pi\epsilon V / \ln(a/b)$ on the inner cylinder and $\lambda_o = +2\pi\epsilon V / \ln(a/b)$ on the outer cylinder, and field $\mathbf{E} = -V / [r\,\ln(a/b)]\,\hat{\mathbf{e}}_r$.
\end{tcolorbox}

\begin{tcolorbox}[colframe=black,colback=white,title=Qwen3.5-9B]
``[\ldots] the potential difference $V$ is the line integral of the electric field from the inner cylinder (radius~$a$) to the outer cylinder (radius~$b$).''

Final answer: $\lambda_i = -2\pi\epsilon V / \ln(b/a)$; \quad $\mathbf{E} = -V / [r\,\ln(b/a)]\,\hat{\mathbf{r}}$.

\begin{tcolorbox}[colback=red!10, colframe=red!80, boxrule=0.75pt]
\textbf{Error:} The figure labels the inner radius~$b$ and the outer radius~$a$. The model reads them in the conventional order ($a$ inner, $b$ outer), reversing the logarithm's argument. Since $\ln(b/a) = -\ln(a/b)$, both the charge signs and the field direction are wrong.
\end{tcolorbox}
\end{tcolorbox}

\begin{tcolorbox}[colframe=black,colback=lightred,title=Error Category: Figure Misreading]
\textbf{Error reason:} The model assumes the standard convention for radius labels instead of reading them from the diagram, reversing the charge assignment and field direction.
\end{tcolorbox}

%% file: figures_tex/calculation_error.tex
\begin{tcolorbox}[colframe=black,colback=lightgrey]
\textbf{Question:}

A thin uniform bar of mass $m$ and length $\frac{3l}{2}$ is suspended by a string of length $l$ and negligible mass. Give the normal frequencies and normal modes for small oscillations in a plane.\\

\textbf{Reference answer:}\smallskip

$\omega_1 = (\sqrt{3} + 1) \sqrt{\frac{g}{l}}, \quad \omega_2 = (\sqrt{3} - 1) \sqrt{\frac{g}{l}}$.
\end{tcolorbox}

\begin{tcolorbox}[colframe=black,colback=white,title=Qwen3.5-9B]
The model correctly sets up the center-of-mass and rotational kinetic energies for the string--bar system but writes:

``This led to $M_{12} = 3/8\, ml^2$.''

The off-diagonal mass-matrix coefficient should be $3/4\, ml^2$; the model introduces an erroneous factor of $1/2$.

Final answer: $\omega_1 = \sqrt{\frac{4g}{l(4+\sqrt{3})}}, \quad \omega_2 = \sqrt{\frac{4g}{l(4-\sqrt{3})}}$.

\begin{tcolorbox}[colback=red!10, colframe=red!80, boxrule=0.75pt]
\textbf{Error:} Halving the off-diagonal element changes the characteristic equation and both normal frequencies.
\end{tcolorbox}
\end{tcolorbox}

\begin{tcolorbox}[colframe=black,colback=lightred,title=Error Category: Calculation Errors]
\textbf{Error reason:} The physical setup is correct, but a factor-of-two algebraic error propagates into both eigenfrequencies.
\end{tcolorbox}

%% file: figures_tex/assumption.tex
\begin{tcolorbox}[colframe=black,colback=lightgrey]
\textbf{Question:}

A point source $S$ of monochromatic light emits radiation of frequency $f$. An observer $A$ moves at constant speed $v$ along a straight line that passes at a distance $d$ from the source.

(a) Derive an expression for the observed frequency as a function of the distance $x$ from the point of closest approach.\\

\textbf{Reference answer:}\smallskip

$f' = \left(1 - \frac{\beta x}{\sqrt{d^2 + x^2}} \right)\frac{f}{\sqrt{1-\beta^2}}$, \quad where $\beta = v/c$.
\end{tcolorbox}

\begin{tcolorbox}[colframe=black,colback=white,title=Qwen3.5-9B]
``The exact formula would involve the Lorentz factor, typically resulting in $f' = f \sqrt{\frac{1 - v_r/c}{1 + v_r/c}}$ [\ldots] However, given the likely context (Chapter~3), the classical approximation is the standard expected answer.''

Final answer: $f' = f \left( 1 - \frac{vx}{c\sqrt{x^2 + d^2}} \right)$.

\begin{tcolorbox}[colback=red!10, colframe=red!80, boxrule=0.75pt]
\textbf{Error:} The model recognizes the relativistic formula but drops the Lorentz factor based on a guess about the textbook chapter. No low-speed limit is stated in the problem, so omitting the $1/\sqrt{1-\beta^2}$ factor changes the observed frequency.
\end{tcolorbox}
\end{tcolorbox}

\begin{tcolorbox}[colframe=black,colback=lightred,title=Error Category: Non-Existent Assumptions]
\textbf{Error reason:} The model introduces a nonrelativistic approximation not warranted by the problem statement.
\end{tcolorbox}

%% file: figures_tex/misunderstanding.tex
\begin{tcolorbox}[colframe=black,colback=lightgrey]
\textbf{Question:}

A box contains a mixture of ideal gas and some special molecules which can adsorb the gas. Each molecule contains two sites at which it can adsorb a single ideal gas particle in either one of two states: one with binding energy $-\epsilon_0$ and another with $-\epsilon_0/2$. [\ldots] Determine the fraction of the molecules with exactly one adsorption site full, as a function of the gas pressure and temperature.\\

\textbf{Reference answer:}\smallskip

$P_{2}^{1} = \dfrac{2p_0\, p \left(1 + e^{-\epsilon_0 / (2 \tau)}\right)}{\left[p_0 + p \left(1 + e^{-\epsilon_0 / (2 \tau)}\right)\right]^2}$
\end{tcolorbox}

\begin{tcolorbox}[colframe=black,colback=white,title=Qwen3.5-9B]
``Site~1 has a binding energy of $-\epsilon_0$. [\ldots] Site~2 has a binding energy of $-\epsilon_0/2$.''

The model treats the two binding energies as belonging to two different sites rather than as two possible states at each site. This changes the single-site partition function and the resulting probability.

\begin{tcolorbox}[colback=red!10, colframe=red!80, boxrule=0.75pt]
\textbf{Error:} The question states that each site admits either binding state. Assigning one energy per site reduces the state count and produces the wrong partition function.
\end{tcolorbox}
\end{tcolorbox}

\begin{tcolorbox}[colframe=black,colback=lightred,title=Error Category: Misunderstanding the Question]
\textbf{Error reason:} The model misinterprets the problem's state structure, answering a simpler problem than the one posed.
\end{tcolorbox}